\documentclass[letterpaper, 10 pt, conference]{ieeeconf}
\IEEEoverridecommandlockouts
\pdfoutput=1

\let\labelindent\relax
\usepackage{graphicx} \graphicspath{ {figures/} }
\usepackage{amsmath,amssymb,mathabx}
\usepackage[cal=cm]{mathalfa}
\usepackage{mathtools}
\usepackage{algorithmic}
\usepackage[ruled]{algorithm2e}
\usepackage{acronym}
\usepackage{enumitem}
\usepackage{booktabs}
\usepackage{hyperref}
\usepackage[stable]{footmisc}
\usepackage{balance}
\usepackage{xspace,setspace}
\usepackage[skip=3pt,font=small]{subcaption}
\usepackage[skip=3pt,font=small]{caption}
\usepackage[misc]{ifsym}
\usepackage[dvipsnames]{xcolor}
\usepackage[capitalise]{cleveref}
\usepackage{tabularx,colortbl,multirow,array,makecell}
\usepackage{overpic}
\usepackage{cite}

\definecolor{Gray}{gray}{0.85}
\newcolumntype{a}{>{\columncolor{Gray}}c}

\makeatletter
\DeclareRobustCommand\onedot{\futurelet\@let@token\@onedot}
\def\@onedot{\ifx\@let@token.\else.\null\fi\xspace}
\def\eg{\emph{e.g}\onedot} 
\def\ie{\emph{i.e}\onedot}

\def\etal{\emph{et al}\onedot}
\makeatother

\crefname{algocf}{alg.}{algs.}
\Crefname{algocf}{Algorithm}{Algorithms}

\makeatletter
\def\BState{\State\hskip-\ALG@thistlm}
\makeatother

\makeatletter
\renewcommand{\paragraph}{%
  \@startsection{paragraph}{4}%
  {\z@}{0ex \@plus 0ex \@minus 0ex}{-1em}%
  {\hskip\parindent\normalfont\normalsize\bfseries}%
}
\makeatother

\crefname{algocf}{alg.}{algs.}
\Crefname{algocf}{Algorithm}{Algorithms}

\definecolor{gblue}{HTML}{4285F4}
\definecolor{gred}{HTML}{DB4437}

\acrodef{mpc}[MPC]{Model Predictive Control}
\acrodef{slam}[SLAM]{Simultaneous Localization and Mapping}
\acrodef{sfm}[SfM]{Structure from Motion}
\acrodef{tamp}[TAMP]{Task and Motion Planning}
\acrodef{urdf}[URDF]{Unified Robot Description Format}
\acrodef{pg}[\emph{pg}]{parse graph}
\acrodef{pt}[\emph{pt}]{parse tree}
\acrodef{me}[ME]{Matching Error}
\acrodef{ae}[AE]{Alignment Error}

\title{\LARGE \bf
Reconstructing Interactive 3D Scenes\\by Panoptic Mapping and CAD Model Alignments%
}%

\author{Muzhi Han$^{*}$\quad{}Zeyu Zhang$^{*}$\quad{}Ziyuan Jiao\quad{}Xu Xie\quad{}Yixin Zhu\quad{}Song-Chun Zhu\quad{}Hangxin Liu\quad{}
\thanks{$^{*}$ Muzhi Han and Zeyu Zhang contributed equally to this work.}%
\thanks{UCLA Center for Vision, Cognition, Learning, and Autonomy (VCLA) at the Statistics Department. Emails:
\{muzhihan, zeyuzhang, zyjiao, xiexu, yixin.zhu, hx.liu\}@ucla.edu, sczhu@stat.ucla.edu.}%
\thanks{The work reported herein was supported by ONR N00014-19-1-2153, ONR MURI N00014-16-1-2007, and DARPA XAI N66001-17-2-4029.}%
}

\begin{document}

\maketitle
\thispagestyle{empty}
\pagestyle{empty}

\begin{abstract}
In this paper, we rethink the problem of scene reconstruction from an embodied agent's perspective: While the classic view focuses on the reconstruction accuracy, our new perspective emphasizes the underlying functions and constraints such that the reconstructed scenes provide \emph{actionable} information for simulating \emph{interactions} with agents. Here, we address this challenging problem by reconstructing an interactive scene using RGB-D data stream, which captures (i) the semantics and geometry of objects and layouts by a 3D volumetric panoptic mapping module, and (ii) object affordance and contextual relations by reasoning over physical common sense among objects, organized by a graph-based scene representation. Crucially, this reconstructed scene replaces the object meshes in the dense panoptic map with part-based articulated CAD models for finer-grained robot interactions. In the experiments, we demonstrate that (i) our panoptic mapping module outperforms previous state-of-the-art methods, (ii) a high-performant physical reasoning procedure that matches, aligns, and replaces objects' meshes with best-fitted CAD models, and (iii) reconstructed scenes are physically plausible and naturally afford actionable interactions; without any manual labeling, they are seamlessly imported to ROS-based simulators and virtual environments for complex robot task executions.\footnote{The code is available at \url{https://github.com/hmz-15/Interactive-Scene-Reconstruction}.}
\end{abstract}

\setstretch{0.995}

\section{Introduction}

Perception of the human-made scenes and the objects within inevitably leads to the course of actions~\cite{gibson1950perception,gibson1966senses}; such a task-oriented view~\cite{ikeuchi1992task,zhu2015understanding} is the basis for a robot to interact with the environment and accomplish complex tasks. In stark contrast, such a crucial perspective is largely missing in the robot mapping and scene reconstruction literature: Prevailing semantic mapping or \ac{slam} methods often produce a metric map of the scene with semantic or instance annotations; they only emphasize mapping accuracy but omit the essence of robot task execution---actions that a semantic entity could afford and associated physical constraints embedded among entities. 

\begin{figure}[t!]
    \centering
    \includegraphics[width=\linewidth]{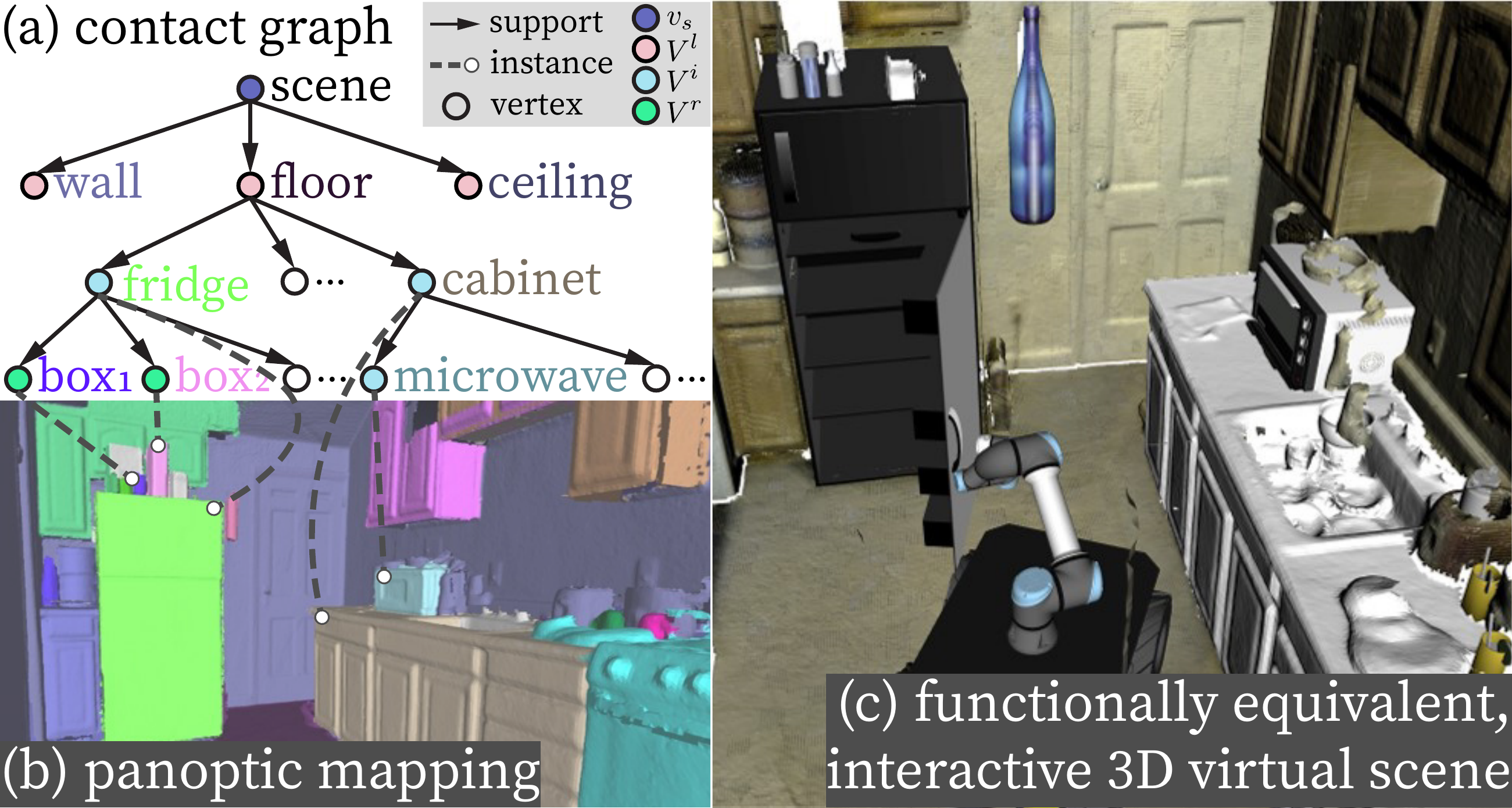}
    \caption{\textbf{The reconstruction of an interactive 3D scene.} (a) A contact graph is constructed by the supporting relations that emerged from (b) panoptic mapping. By reasoning their affordance, functional objects within the scene are matched and aligned with part-based interactive CAD models. (c) The reconstructed scene enables a robot simulates its task execution with comparable outcomes in the physical world.}
    \label{fig:motivation}
\end{figure}

Such a lack of the scene's functional representation leads to a gap between the reconstructed semantic scenes and \ac{tamp}, which prevents a robot from directly interacting with the reconstructed scenes to accomplish complex tasks. Take the reconstructed scene in \cref{fig:motivation} as the example, wherein the robot is tasked to pick up a frozen meal from the fridge, microwave and serve it. To properly plan and execute inside the reconstructed scene, a robot ought to acquire (i) semantics and geometry of objects (\eg, this piece of point cloud is a fridge), (ii) actions an object affords (\eg, a fridge can be open), and (iii) constraints among these entities (\eg, no objects should float in the air). Although modern semantic mapping and \ac{slam} methods can partially address (i)~\cite{hoang2020panoptic,narita2019panopticfusion}, existing solutions for (ii)~\cite{myers2015affordance,zhu2015understanding,zhu2016inferring} and (iii)~\cite{zheng2013beyond,zheng2014detecting,zhao2013scene,zheng2015scene,huang2018cooperative,chen2019holistic++} have not yet been fully integrated into a robot scene reconstruction framework, resulting in non-interactive reconstructed scenes. This deficiency precludes the feasibility of directly applying \ac{tamp} on the reconstructed scenes either using traditional~\cite{kaelbling2011hierarchical,srivastava2014combined} or learning-based~\cite{kim2019learning,wang2018active} methods; the robot can hardly verify whether its plan is valid or the potential outcomes of its actions are satisfied before executing in the physical world.

Although researchers have attempted to devise manual pipelines (\eg, iGibson~\cite{xia2020interactive}, SAPIEN~\cite{xiang2020sapien}) to either convert the reconstructed real-world scenes or directly build virtual environments from scratch, creating such simulation environments is a non-trivial and time-consuming task. The simulated environment should be sufficiently similar to the reality, and the objects to be interacted with should afford sufficiently similar functionality. Only by satisfying the above conditions could the outcomes of interactions in simulation be similar to those in the physical world. Due to the enormous workload to create/convert each scene, the number of available scenes to date is still quite limited. A challenge naturally arises: Can we reconstruct a scene that can be automatically imported into various simulators for interactions and task executions?

In this paper, we propose a new task of reconstructing \emph{functionally equivalent} and interactive scenes, capable of being directly imported into simulators for robot training and testing of complex task execution. We argue that a scene's functionality is composed of the functions afforded by objects within the scene. Therefore, the essence of our scene reconstruction lies in defining functionally equivalent objects, which should preserve four characteristics with decreasing importance: (i) its semantic class and spatial relations with nearby objects, (ii) its affordance, \eg, what interactions it offers, (iii) a similar geometry in terms of size and shape, and (iv) a similar appearance.

Existing approaches oftentimes represent reconstructed semantic scene and its entities as sparse landmarks~\cite{pronobis2012large,yang2019cubeslam}, surfels~\cite{mccormac2017semanticfusion,hoang2020panoptic}, or volumetric voxels~\cite{grinvald2019volumetric,mccormac2018fusion++}. However, these representations are inadequate to serve as a \emph{functional representation} of the scene and its entities: They merely provide occupancy information (\ie, where the fridge is) without any actionable information for robot interactions or planning (\eg, whether or how the fridge can be open).

\begin{figure*}[t!]
    \centering
    \includegraphics[width=\linewidth]{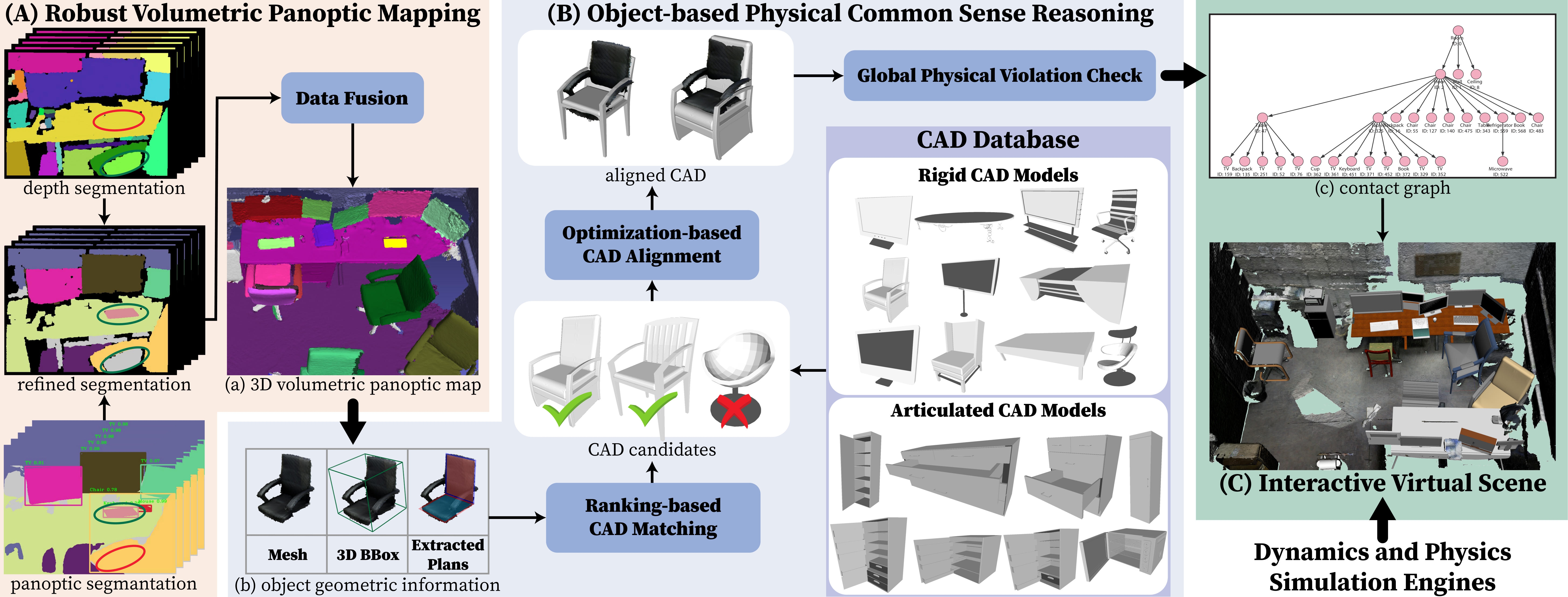}
    \caption{\textbf{System architecture for reconstructing a functionally equivalent scene.} (A) Per-frame segmentation and cross-frame data fusion produce (a) a 3D volumetric panoptic map with fine-grained semantics and geometry, served as the input for (B) physical common sense reasoning that matches, aligns, and replaces segmented object meshes with functionally equivalent CAD alternatives. Specifically, (b) by geometric similarity, a ranking-based matching algorithm selects a shortlist of CAD candidates, followed by an optimization-based process that finds a proper transformation and scaling between the CAD candidates and object mesh. A global physical violation check is further applied to finalize CAD replacements to ensure physical plausibility. (C) This CAD augmented scene can be seamlessly imported to existing simulators; (c) contact graph encodes the kinematic relations among the objects in the scene as the planning space for a robot.}
    \label{fig:sys_arch}
\end{figure*}

To address the above issues, we devise three primary components in our system; see an illustration in \cref{fig:sys_arch}:

\textbf{(A) A robust 3D volumetric panoptic mapping module}, detailed in \cref{sec:mapping}, accurately segments and reconstructs 3D objects and layouts in clustered scenes even with noisy per-frame image segmentation results. The term ``panoptic,'' introduced in~\cite{kirillov2019panoptic}, refers to jointly segmenting \emph{stuff} and \emph{things}. In this paper, we regard objects as \emph{things} and layout as \emph{stuff}. Our system produces a volumetric panoptic map using a novel per-frame panoptic fusion and a customized data fusion procedure; see examples in \cref{fig:motivation}b and \cref{fig:sys_arch}a.

\textbf{(B) A physical common sense reasoning module}, detailed in \cref{sec:reasoning}, replaces object meshes obtained from the panoptic map with interactive rigid or articulated CAD models. This step is achieved by a ranking-based CAD matching and an optimization-based CAD alignment, which accounts for both geometric and physical constraints. We further introduce a global physical violation check to ensure that every CAD replacement is physically plausible.

\textbf{(C) A graphical representation}, \emph{contact graph} $cg$, (\cref{fig:motivation}a, \cref{fig:sys_arch}c, and \cref{sec:represent}) is built and maintained simultaneously, in which the nodes of a $cg$ represent objects and layouts, and the edges of a $cg$ denote the support and proximal relations. We further develop an interface to convert a $cg$ to a \ac{urdf} such that the reconstructed functionally equivalent scene (see \cref{fig:motivation}C) can be directly imported into simulators for robot interactions and task executions; see \cref{sec:result} for experimental results.

\paragraph*{Related Work}

Existing approaches to generate simulated interactive environments fall into three categories: (i) \textbf{manual efforts}, such as those in Gazebo~\cite{koenig2004design} and V-REP~\cite{rohmer2013v} for robotics, AI2THOR~\cite{kolve2017ai2} and Gibson~\cite{xia2018gibson} for embodied AI, and iGibson~\cite{xia2020interactive}, SAPIEN~\cite{xiang2020sapien}, and VRGym~\cite{xie2019vrgym} with part-based articulated objects (\eg, a cabinet with a door); (ii) \textbf{scene synthesis} that produces a massive amounts scenes with the help of CAD databases~\cite{yu2011make,qi2018human,jiang2018configurable}; (iii) \textbf{large-scale scene dataset} with aligned CAD models, such as SUNCG~\cite{song2017semantic} and 3D-FRONT~\cite{fu20203dfront}. However, without tedious manual work, all of these prior approaches fail to replicate a real scene in simulation with diverse interactions.

\setstretch{0.97}

Modern \textbf{semantic mapping}~\cite{narita2019panopticfusion,grinvald2019volumetric,pham2019real} and \textbf{object \ac{slam}}~\cite{yang2019cubeslam,mccormac2018fusion++} methods can effectively reconstruct an indoor scene at an object-level. Physical cues, such as support and collision, have been further integrated to estimate and refine the object pose~\cite{yang2019monocular,wada2020morefusion,sui2020geofusion}. In parallel, computer vision algorithms predict \textbf{3D instance segmentation} in densely reconstructed scenes~\cite{yi2019gspn,pham2019jsis3d}, and then fit CAD models by crowdsourcing~\cite{dai2017scannet} or by computing the correspondences between the reconstructed scenes and CAD models~\cite{avetisyan2019scan2cad,avetisyan2020scenecad}. However, the above work fails to go beyond semantics to (i) capture the interactive nature of the objects, or (ii) meaningfully represent a physically plausible scene. As such, the reconstructed scenes still fail to be imported into simulators to afford robot interactions and task executions.

Constructing a proper scene or a map \textbf{representation} remains an open problem~\cite{cadena2016past}. Typical semantic mapping and \ac{slam} methods only output a flat representation, difficult to store or process high-level semantics for robot interactions and task executions. Meanwhile, graph-based representations, \eg, scene grammar~\cite{zhu2007stochastic,zhao2011image,huang2018cooperative,zhao2013scene,jiang2018configurable,chen2019holistic++} and 3D scene graph~\cite{armeni20193d,wald2020learning,rosinol20203d}, provide structural and contextual information. In particular, Rosinol \etal~\cite{rosinol20203d} also incorporate actionable information for robot navigation tasks. Our work devises a \emph{contact graph} with supporting and proximal relations, which imposes kinematic constraints for more complex robot manipulation.

\section{Contact-based Scene Representation}\label{sec:represent}

We devise a graph-based representation, \emph{contact graph} $cg$, to represent a 3D indoor scene. Formally, a contact graph $cg = (pt, E)$ contains (i) a \acf{pt} that captures the hierarchical relations among the scene entities~\cite{zhu2007stochastic}, and (ii) the proximal relations $E$ among entities represented by undirected edges; see an example of \ac{pt} in \cref{fig:motivation}a.

\subsection{Representation}

\textbf{Scene Parse Tree}
$pt = (V, S)$ has been used to represent the hierarchical decompositional relations (\ie, the edge set $S$) among entities (\ie, the node set $V$) in various task domains, including 2D images and 3D scenes~\cite{zhao2011image,zhao2013scene,qi2018human,jiang2018configurable,huang2018holistic,huang2018cooperative,chen2019holistic++}, videos and activities~\cite{zhu2015understanding,zhu2016inferring,qi2020generalized}, robot manipulations~\cite{edmonds2017feeling,liu2018interactive,edmonds2019tale,liu2019mirroring,zhang2020graph}, and theory of mind~\cite{yuan2020joint}. In this paper, we adopt \ac{pt} to represent supporting relations among entities, dynamically built and maintained during the reconstruction; for instance in \cref{fig:motivation}a, the \texttt{cabinet} is the parent node of the \texttt{microwave}. Supporting relation is quintessential in scene understanding with physical common sense as it reflects the omnipresent physical plausibility; \ie, if the cabinet were moved, the microwave would move together with it or fall onto the ground. This counterfactual perspective goes beyond occupancy information (\ie, the physical location of an object); in effect, it further provides actionable information and the potential outcome of actions for robot interactions and task executions in the scene.

\textbf{Scene Entity Nodes}
$V=\{v_s\} \cup V^L \cup V^R \cup V^A$ include: (i) the scene node $v_s$, severing as the root of $pt$, (ii) layout node set $V^L$, including floor, ceiling, and the wall that bound the 3D scene, (iii) rigid object set $V^R$, wherein each object has no articulated part (\eg, a table), and (iv) articulated object set $V^A$, wherein each object has articulated parts to be interacted for various robot tasks (\eg, fridge, microwave).
Each non-root node $v_i = \langle o_i, c_i, M_i, B_i(\boldsymbol{p}_i, \boldsymbol{q}_i, \boldsymbol{s}_i), \Pi_i \rangle$ encodes a unique instance label $o_i$, a semantic label $c_i$, a full geometry model $M_i$ (a triangular mesh or a CAD model), a 3D bounding box $B_i$ (parameterized by its position $\boldsymbol{p}_i$, orientation $\boldsymbol{q}_i$, and size $\boldsymbol{s}_i$, all in $\mathbb{R}^3$), and a set of surface planes $\Pi_i=\{\boldsymbol{\pi}_i^k, k = 1 \cdots |\Pi_i|\}$, where $\boldsymbol{\pi}_i^k$ is a homogeneous vector $[{\boldsymbol{n}_i^k}^T, d_i^k]^T \in \mathbb{R}^4$ in the projective space~\cite{hartley2003multiple} with unit plane normal vector $\boldsymbol{n}_i^k$, and any point $\boldsymbol{v} \in \mathbb{R}^3$ on the plane satisfies a constraint: ${\boldsymbol{n}_i^k}^T \cdot \boldsymbol{v} + d_i^k=0$.

\textbf{Supporting Relations}
$S$ is the set of directed edges in $pt$ from parent nodes to their child nodes. Each edge $s_{p,c} \in S$ imposes physical common sense between the parent node $v_p$ and the child node $v_c$. These constraints are necessary to ensure that $v_p$ supports $v_c$ in a physically plausible fashion:

\noindent\textbf{(i) Geometrical plausibility:} $v_p$ should have a plane $\boldsymbol{\pi}_p^s=[{\boldsymbol{n}_p^s}^T, d_p^s]^T$ with $\boldsymbol{n}_p^s$ being opposite to the gravity direction, whereas bottom surface of $v_c$ should contact the top of $\boldsymbol{\pi}_p^s$:
\begin{equation}
\resizebox{0.91\linewidth}{!}{%
    $\displaystyle
    \exists \boldsymbol{\pi}_p^s \in \Pi_p, {\boldsymbol{n}_p^s}^T \cdot \boldsymbol{g}\leqslant a_{th},
    ~s.t.~\mathcal{D}(v_c, \boldsymbol{\pi}_p^s) = p_c^g - (-d_p^s + s_c^g / 2)=0,$
    \label{equ:supportingplane}
}%
\end{equation}
where $\boldsymbol{g}$ is the unit vector along the gravity direction, $a_{th}=-0.9$ is a tolerance coefficient, $d_p^s$ is the offset of the $v_p$'s supporting plane, and $p_c^g$ and $s_c^g$ denote the position and size of the $v_c$'s 3D bounding box along the gravity direction.

\noindent\textbf{(ii) Sufficient contact area for stable support:} Formally,
\begin{equation}
    \mathcal{A}(v_p, v_c) = \text{A}(v_p \cap v_c) / \text{A}(v_c) \geqslant b_{th},
    \label{equ:supportingarea}
\end{equation}
where $\text{A}(v_c)$ is the bottom surface of the $v_c$'s 3D bounding box, and $\text{A}(v_p \cap v_c)$ is the area of the overlapping rectangle containing the mesh vertices of $v_p$ near $\boldsymbol{\pi}_p^s$ within $v_c$'s 3D bounding box. We set threshold $b_{th}=0.5$ for a stable support.

\textbf{Proximal Relations}
$E$ introduce links among entities in the \ac{pt}. They impose additional constraints by modeling spatial relations between two non-supporting but physically nearby objects $v_1$ and $v_2$: Their meshes should not penetrate with each other, \ie, $\text{Vol}(M_1 \cap M_2) = 0$. Note that the constraint only exists between two objects with overlapping 3D bounding boxes, \ie, when $\text{Vol}(B_1 \cap B_2) > 0$.

\subsection{Constructing Contact Graph}\label{sec:graph_init}

Each node $v_x$ in $cg$ is constructed from a scene entity $x$ in the panoptic map (see \cref{sec:mapping}) by: (i) acquiring its $o_x, c_x, M_x, B_x(\boldsymbol{p}_x, \boldsymbol{q}_x, \boldsymbol{s}_x)$, (ii) extracting surface planes $\Pi_x$ by iteratively applying RANSAC~\cite{taguchi2013point} and removing plane inliers, and (iii) assigning $x$ as $v_x$ in $cg$.

Given a set of nodes constructed on-the-fly, we apply a bottom-up process to build up $cg$ by detecting supporting relations among the entities. Specifically, given an entity $v_c$, we consider all entities $\{v_i\}$ whose 3D bounding boxes are spatially below it and have proper supporting planes $\boldsymbol{\pi}_i^k$ based on \cref{equ:supportingplane}. The most likely supporting relation is chosen by maximizing the following score function:
\begin{equation}
\resizebox{0.91\linewidth}{!}{%
    $\displaystyle 
    S(v_c,v_i,\boldsymbol{\pi}_i^k) = \left\{1 - \min{\left[1, \|\mathcal{D}(v_c, \boldsymbol{\pi}_i^k)\|\right]}\right\}
    \times \mathcal{A}(v_i, v_c)$,
}%
\end{equation}
where the first term indicates the alignment between the $v_c$'s bottom surface and the $v_i$'s supporting planes, and the second term reflects an effective supporting area, both normalized to $[0, 1]$. $B_i$ is further refined (see \cref{equ:supportingplane}) as it was computed based on incomplete object meshes. Meanwhile, the proximal relations are assembled by objects' pairwise comparison. At length, the $cg$ of the scene is constructed based on the identified entities and their relations and grows on-the-fly.

\section{Robust Panoptic Mapping}\label{sec:mapping}

Robust and accurate mapping of scene entities within clustered environments is essential for constructing a $cg$ and serving downstream tasks. Below, we describe our robust panoptic mapping module to generate volumetric object and layout segments in the form of meshes from RGB-D streams; see the pipeline in \cref{fig:sys_arch}A. We follow the framework proposed in~\cite{grinvald2019volumetric} and \textbf{only highlight crucial technical modifications below}. The experiments demonstrate that our modifications significantly improve system performance.

\paragraph*{Per-frame Segmentation}

We combine the segmentation of both RGB and depth for performance improvement as in~\cite{grinvald2019volumetric}. However, instead of merely labeling the depth segments with semantic-instance masks, we bilaterally fuse panoptic masks and geometric segments to output point cloud segments with both semantic and instance labels. We further perform an outlier removal for each object entity; far away segments are removed and assigned to the scene background.

This modification significantly improves the noisy per-frame segmentation; see \cref{fig:sys_arch}a. In this example, fusing RGB and depth segments mutually improves the segments if they were obtained by each alone. The fusion (i) correctly segments the keyboard and divides the two monitors when depth segments fail, and (ii) geometrically refines the noisy panoptic mask of the chair to exclude the far-away ground.

\paragraph*{Data Fusion}

Compared to~\cite{grinvald2019volumetric}, we introduce two notable enhancements in data fusion. First, we use a triplet count $\Phi(l,c,o)$ to record the frequency that an instance label $o$, a semantic label $c$, and a geometric label $l$ associated with the same point cloud segment; it is incrementally updated: $\Phi(l,c,o)=\Phi(l,c,o)+1$. This modification improves consistency in semantic-instance fusion. Second, in addition to merging two geometric labels if they share voxels over a certain ratio, we also regulate two instance labels if the duration of association with a common geometric label exceeds a threshold. We further estimate a gravity-aligned, 3D-oriented bounding box for each object mesh~\cite{malandain2002computing}. In sum, our system simultaneously and comprehensively outputs a set of scene entities with their instance labels, semantic labels, 3D bounding boxes, and reconstructed meshes.

\paragraph*{Implementation and Evaluation} 

We use an off-the-shelf panoptic segmentation model~\cite{wu2019detectron2} pre-trained on the COCO panoptic class~\cite{lin2014microsoft} for RGB images and a geometric segmentation method~\cite{furrer2018incremental} for depth images. We compare our panoptic mapping module with the original Voxblox++~\cite{grinvald2019volumetric} on 8 sequences in the SceneNN dataset~\cite{hua2016scenenn}. Our evaluation includes four criteria: (i) panoptic quality (PQ)~\cite{kirillov2019panoptic,narita2019panopticfusion}, (ii) segmentation quality (SQ), (iii) recognition quality (RQ) of 3D panoptic mapping on 8 \emph{thing} classes and 2 \emph{stuff} classes, and (iv) the mean average precision (mAP) computed using an intersection of union (IoU) with a threshold of $0.5$ for 3D oriented bounding box estimation on \emph{thing} classes. Since the supporting relations in $cg$ could further refine the 3D bounding boxes (see \cref{sec:graph_init}), we also include mAP$_{\texttt{re}}$.

\cref{tab:map_result} tabulates the class-averaged results, showing that our method consistently outperforms the baseline in both 3D panoptic mapping and 3D bounding box estimation; see \cref{fig:scene_results}b for some qualitative results. In general, refining objects' 3D bounding boxes with supporting relations introduces a significant improvement in accuracy.

\begin{table}[ht!]
    \centering
    \caption{\textbf{Quantitative results of 3D panoptic mapping and 3D oriented bounding box estimation on 8 sequences in the SceneNN dataset~\cite{hua2016scenenn}.}}
    \label{tab:map_result}
    \resizebox{\linewidth}{!}{%
        \begin{tabular}{c | a c a c c | a c a c|}
             \cline{2-10}
             & \multicolumn{5}{c|}{Ours} & \multicolumn{4}{c|}{Voxblox++~\cite{grinvald2019volumetric}} \\
            \cline{2-10}
            Sequence ID& PQ & SQ & RQ & mAP & mAP$_\texttt{re}$ & PQ & SQ & RQ & mAP \\
            \hline
            061 & \textbf{43.0} & 52.0 & \textbf{46.3} & \textbf{33.6} & 33.6 & 25.7 & \textbf{53.1} & 32.2 & 8.9 \\
            086 & \textbf{27.3} & \textbf{39.6} & \textbf{34.6} & \textbf{33.8} & 33.8 & 19.4 & 32.9 & 25.2 & 7.9 \\
            096 & \textbf{12.5} & \textbf{21.4} & \textbf{14.6} & \textbf{14.6} & 14.6 & 7.3 & 11.0 & 8.3 & 14.6 \\
            223 & \textbf{49.5} & \textbf{60.2} & \textbf{63.3} & 24.2 & 55.6 & 21.7 & 40.2 & 26.7 & \textbf{61.4} \\
            225 & \textbf{35.4} & \textbf{46.9} & \textbf{44.8} & \textbf{31.5} & 31.5 & 21.6 & 43.6 & 29.4 & 11.2 \\
            231 & \textbf{37.8} & \textbf{45.9} & \textbf{45.4} & 29.2 & \textbf{31.3} & 17.9 & 30.4 & 22.1 & 19.4 \\
            249 & \textbf{24.4} & 33.8 & \textbf{34.4} & 48.9 & \textbf{71.9} & 23.4 & \textbf{36.4} & 30.6 & 48.5 \\
            322 & \textbf{68.4} & \textbf{71.1} & \textbf{80.0} & 58.3 & \textbf{83.3} & 43.6 & 64.6 & 52.9 & 25.0 \\
            \hline
        \end{tabular}
    }%
\end{table}

\section{Physical Reasoning for CAD Alignments}\label{sec:reasoning}

Due to occlusion or limited camera view, the reconstructed meshes of the scene are oftentimes incomplete. As such, the segmented object meshes are incomplete and non-interactive before recovering them as full 3D models; see examples in \cref{fig:score}a and \cref{fig:alignment}a. We introduce a multi-stage framework to replace a segmented object mesh with a functionally equivalent CAD model. This framework consists of an object-level, coarse-grained CAD matching and fine-grained CAD alignment, followed by a scene-level, global physical violation check; see an illustration in \cref{fig:sys_arch}B.

\subsection{CAD Pre-processing}

We collected a CAD database consisting of both rigid and articulated CAD models, organized by semantic classes. The rigid CAD models are obtained from ShapeNetSem~\cite{chang2015shapenet}, whereas articulated parts are first assembled and then properly transformed into one model. Each CAD is transformed to have its origin and axes aligned with its canonical pose. \cref{fig:sys_arch}B shows some instances of CAD models in the database. Similar to a segmented object entity, a CAD model $y$ is parameterized by $o_y, c_y, M_y, B_y(\boldsymbol{p}_y, \boldsymbol{q}_y, \boldsymbol{s}_y),$ and $\Pi_y$.

\subsection{Ranking-based CAD Matching}

Take the chair in \cref{fig:sys_arch}b as an example: Given a segmented object entity $x$, the algorithm retrieves all CAD models in the same semantic category (\ie, chair) from the CAD database to best fit $x$'s geometric information. Since the exact orientation of $x$ is unknown, we uniformly discretize the orientation space into 24 potential orientations. For each rotated CAD model $y$ that aligned to one of the 24 orientations, the algorithm computes a matching distance:
\begin{equation}
    D(x, y) = \omega_1 \cdot d_s(x, y) + \omega_2 \cdot d_{\pi}(x, y) + \omega_3 \cdot d_b(y),
\end{equation}
where $\omega_1=\omega_2=1.0$ and $\omega_3=0.2$ are the weights of three terms, set empirically. We detail these terms below.

(i) $d_s$ matches the relative sizes of 3D bounding boxes: 
\begin{equation}
    d_s(x, y)=\left\lVert \frac{\boldsymbol{s}_x}{\|\boldsymbol{s}_x\|_2} - \frac{\boldsymbol{s}_y}{\|\boldsymbol{s}_y\|_2}\right\rVert.
\end{equation}

(ii) $d_{\pi}$ penalizes the misalignment between their surface planes in terms plane normal and relative distance:
\begin{equation}
\begin{aligned}
    d_\pi(x, y) =& \min_{f_\Pi} \sum_{\boldsymbol{\pi}_i \in \Pi_x} \left[ \left\lVert \frac{d({T_x}^T\boldsymbol{\pi}_i)}{\|\boldsymbol{s}_x\|_2} - \frac{d(f_\Pi(\boldsymbol{\pi}_i))}{\|\boldsymbol{s}_y\|_2} \right\rVert \right.\\
    & \left. + 1 - \boldsymbol{n}(\boldsymbol{\pi}_i)^T \cdot \boldsymbol{n}(f_{\Pi}(\boldsymbol{\pi}_i)) \right],
\end{aligned}
\end{equation}
where $T_x$ denotes the homogeneous transformation matrix from the map frame on the ground to the frame of the bounding box $B_x$, 
$d(\cdot)$ and $\boldsymbol{n}(\cdot)$ denote the offset and normal vector of a plane, and $f_\Pi: \Pi_x \to \Pi_y$ is a bijection function denoting the assignment of feature planes between $x$ and $y$. Note that $f_\Pi$ is also constrained to preserve supporting planes as defined in \cref{equ:supportingplane}. As computing $d_\pi$ involves solving an optimal assignment problem, we adopt a variant of the Hungarian algorithm~\cite{jonker1987shortest} to identify the best $f_\Pi$. 

(iii) $d_b(y)$ is a bias term that adjusts the overall matching error for less preferable CAD candidates:
\begin{equation}
    d_b(y)=1+\boldsymbol{g}^T \cdot \boldsymbol{z}(y),
\end{equation}
where $\boldsymbol{z}(y)$ denotes the up-direction of the CAD model in the oriented CAD frame, and $\boldsymbol{g}$ is a unit vector along the gravity direction. In general, we prefer CAD candidates that stand upright to those leaning aside or upside down.

\begin{figure}[t!]
    \centering
    \includegraphics[width=\linewidth]{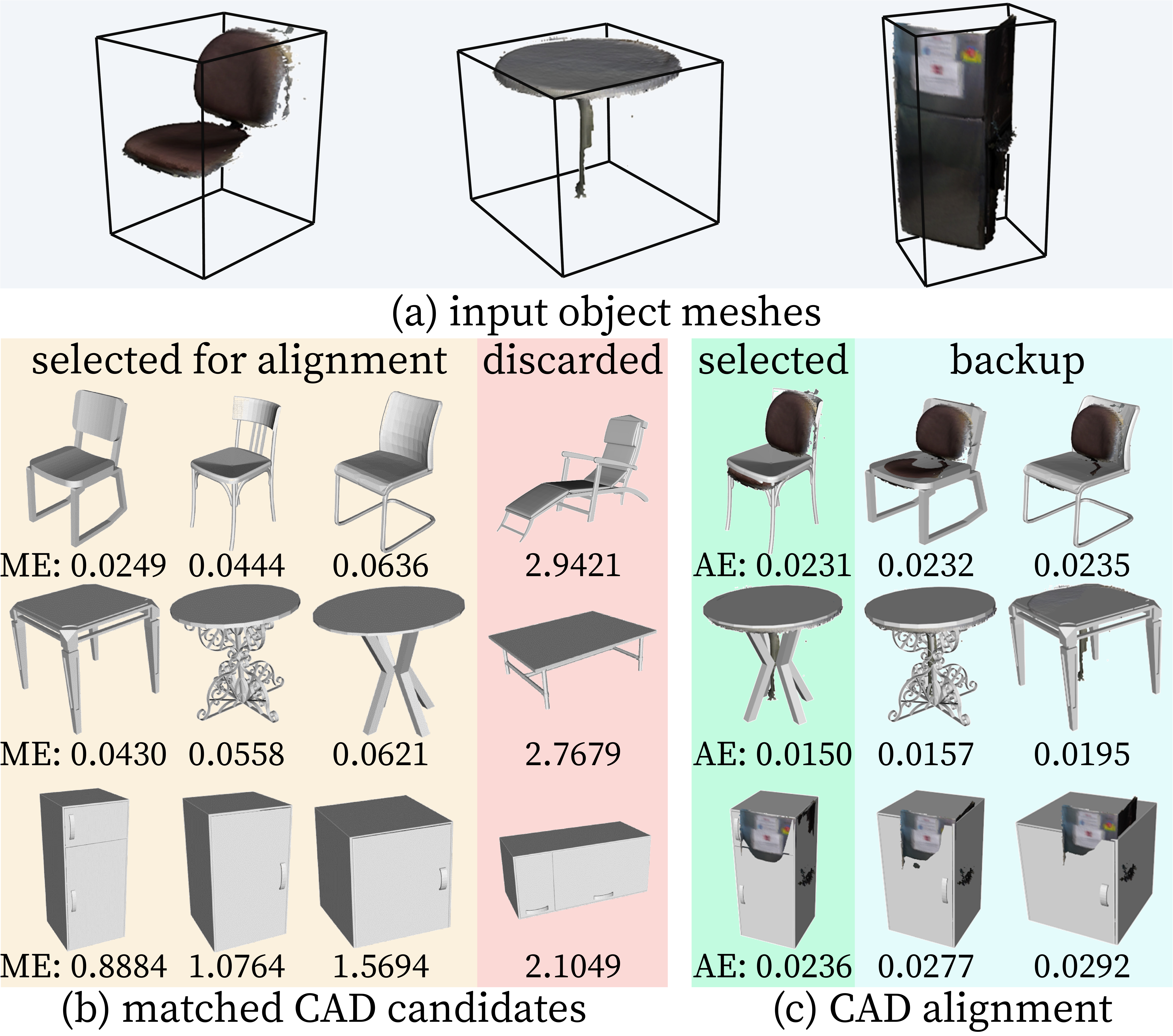}
    \caption{Examples of matching and aligning CAD candidates to (a) an input object mesh. (b) All CAD models within the same semantic class as the input object are retrieved for matching. \acf{me} reflects both the similarity in shapes and the proximity in orientations. After selecting the CAD candidates with smallest \ac{me}s, (c) a fine-grained CAD alignment process selects the best CAD model with a proper transformation based on \acf{ae}.}
    \label{fig:score}
\end{figure}

\cref{fig:score}b illustrates the matching process. Empirically, we observe that the discarded CAD candidates of ``chair'' and ``table'' due to large \acf{me} are indeed more visually distinct from the input object meshes. Moreover, the ``fridge'' model with a wrong orientation has a much larger \ac{me} and is thus discarded. These results demonstrate that our ranking-based matching process can select visually more similar CAD models with the correct orientation. Our system maintains the top $10$ orientated CAD candidates with the lowest \ac{me} for the fine-grained alignment in the next stage.

\begin{figure}[t!]
    \centering
    \includegraphics[width=\linewidth]{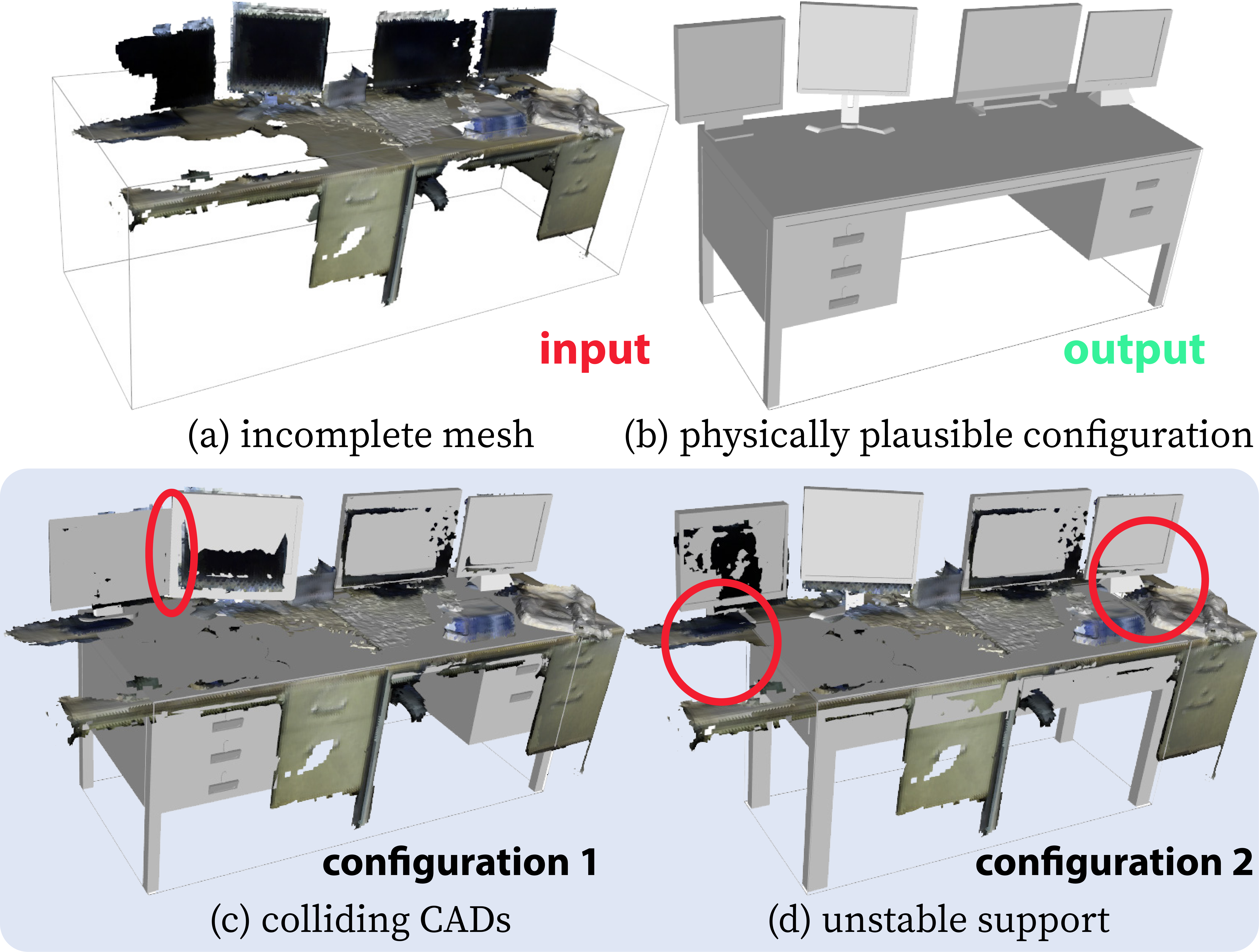}
    \caption{Given (a) incomplete object meshes, our physical common sense reasoning for CAD replacement (b) generates a functionally equivalent and physically plausible configuration. Specifically, the CAD matching and alignment algorithms select and rank a shortlist of CAD candidates. A global physical violation check prunes invalid configurations such as (c) collision and (d) unstable support.}
    \label{fig:alignment}
\end{figure}

\subsection{Optimization-based CAD Alignment}

Given a shortlist of CAD candidates, the overarching goal of this step to find an accurate transformation (instead of 24 discretized orientations) that aligns a given CAD candidate $y$ to the original object entity $x$, achieved by estimating a homogeneous transformation matrix between $x$ and $y$:
\begin{equation}
    T=\left[\begin{matrix}
        \alpha R & \boldsymbol{p}\\
        \boldsymbol{0}^T & 1
    \end{matrix}\right],
    ~s.t.~\min_T{\mathcal{J}(x,T \circ y)},
\end{equation}
where $\circ$ denotes the transformation of a CAD candidate $y$, $\mathcal{J}$ is an alignment error function, $\alpha$ is a scaling factor, $R=Rot(\boldsymbol{z},\theta)$ is a rotation matrix that only considers the yaw angle under the gravity-aligned assumption, and $\boldsymbol{p}$ is a translation. This translation is subject to the following constraint: $p^g=-d^s + \alpha \cdot s^g_y / 2$, as the aligned CAD candidate is supported by a supporting plane $\boldsymbol{\pi}=[{\boldsymbol{n}_\cdot^s}^T,d_\cdot^s]$.

The objective function $\mathcal{J}$ can be written in a least squares form and minimized by the Levenberg–Marquardt~\cite{more1978levenberg} method:
\begin{equation}
    \mathcal{J}=\boldsymbol{e}_b^T \Sigma_b \boldsymbol{e}_b+\boldsymbol{e}_p^T \Sigma_p \boldsymbol{e}_p,
\end{equation}
where $\boldsymbol{e}_b$ is the 3D bounding box error, $\boldsymbol{e}_p$ the plane alignment error, and $\Sigma_b, \Sigma_p$ the error covariance matrices of the error terms. Specifically: (i) $\boldsymbol{e}_b$ aligns the height of the two 3D bounding boxes while constraining the ground-aligned rectangle of the transformed $B_y$ inside that of $B_x$:
\begin{equation}
\small
   \displaystyle \boldsymbol{e}_b=[\text{A}(
   T \circ y))-\text{A}(x,T \circ y),\alpha \cdot \boldsymbol{s}_y^g-\boldsymbol{s}_x^g]^T,
\end{equation}
and (ii) $\boldsymbol{e}_p$ aligns all the matched feature planes as:
\begin{equation}
\fontsize{7.5}{7.5}\selectfont
\begin{aligned}
   \boldsymbol{e}_p&=[\Delta \boldsymbol{\pi}_1,...,\Delta \boldsymbol{\pi}_{|\Pi_x|}]^T, \\
   \displaystyle \Delta \boldsymbol{\pi}_i&=[-d(\boldsymbol{\pi}_i)+ d(T^{-T} \cdot f_\Pi(\boldsymbol{\pi_i})),1- {\boldsymbol{n}(\boldsymbol{\pi}_i})^T \cdot \boldsymbol{n}(T^{-T} \cdot f_{\Pi}(\boldsymbol{\pi}_i))].
\end{aligned}
\end{equation}

We evaluate each aligned CAD candidate by computing an \acf{ae}, the root mean square distance between the object mesh vertices and the closest points on aligned CAD candidate; \cref{fig:score}c shows both qualitative and quantitative results. The CAD candidate with the smallest \ac{ae} will be selected, whereas others are potential substitutions if the selected CADs violate physical constraints, detailed next. 

\begin{figure*}[t!]
    \centering
    \begin{subfigure}[c]{0.2\linewidth}
        \includegraphics[width=\linewidth]{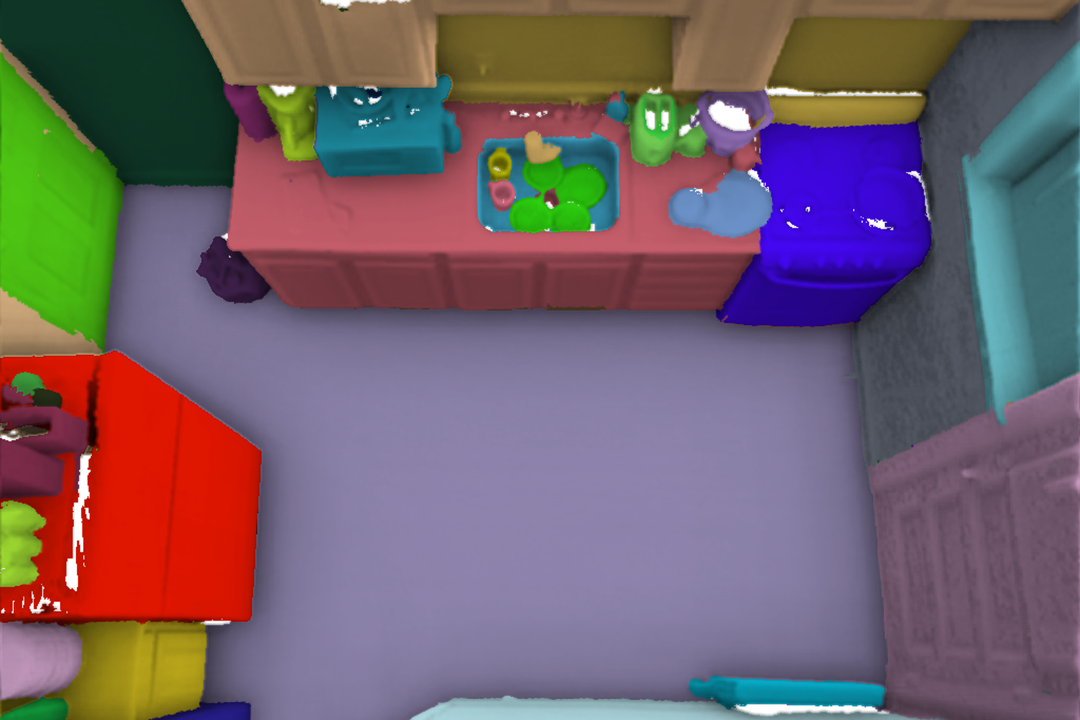}
    \end{subfigure}%
    \begin{subfigure}[c]{0.2\linewidth}
        \includegraphics[width=\linewidth]{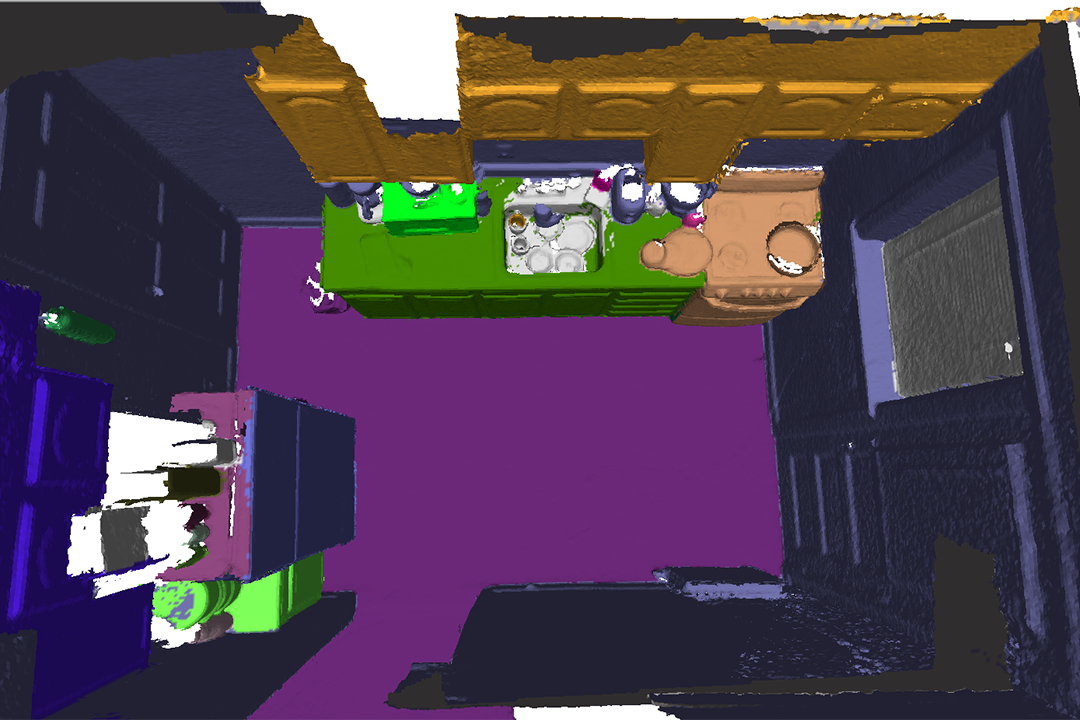}
    \end{subfigure}%
    \begin{subfigure}[c]{0.2\linewidth}
        \includegraphics[width=\linewidth]{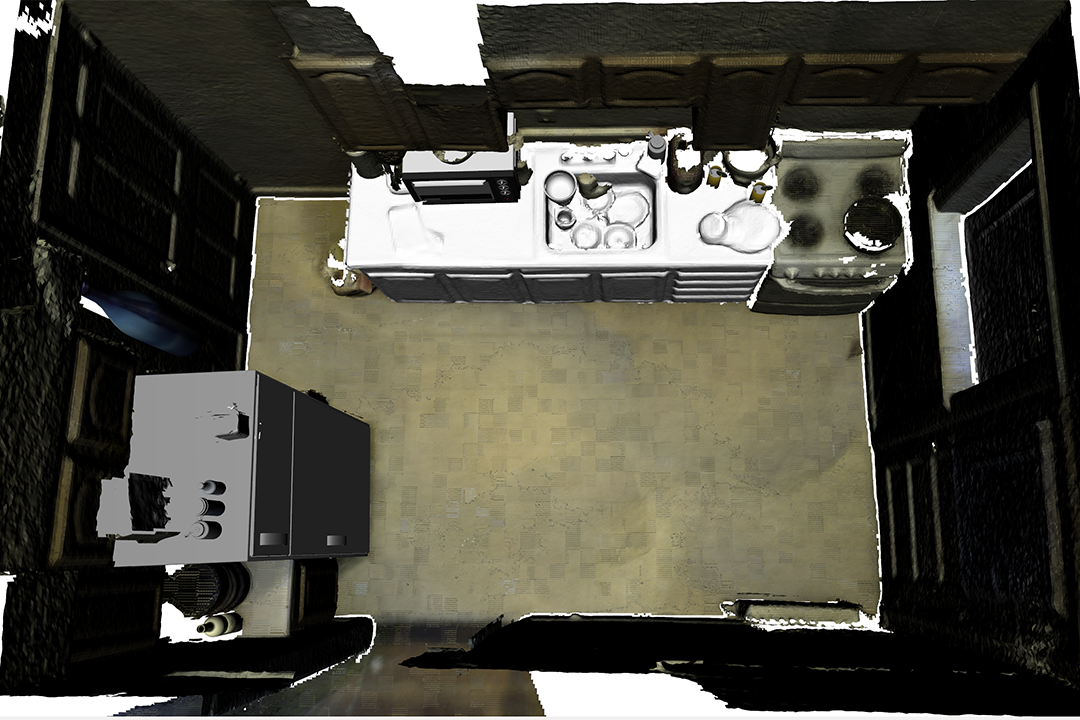}
    \end{subfigure}%
    \begin{subfigure}[c]{0.2\linewidth}
        \begin{overpic}[width=\linewidth]{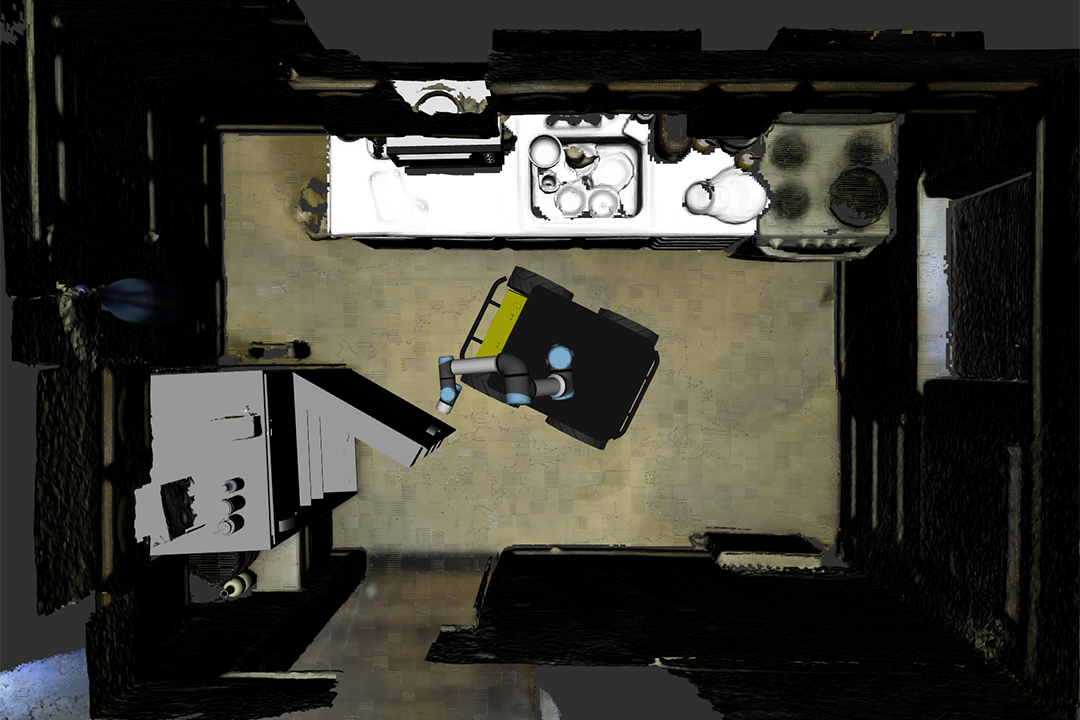}
            \put(68,1){\color{white}\linethickness{0.5mm}%
                \frame{\includegraphics[width=0.3\linewidth]{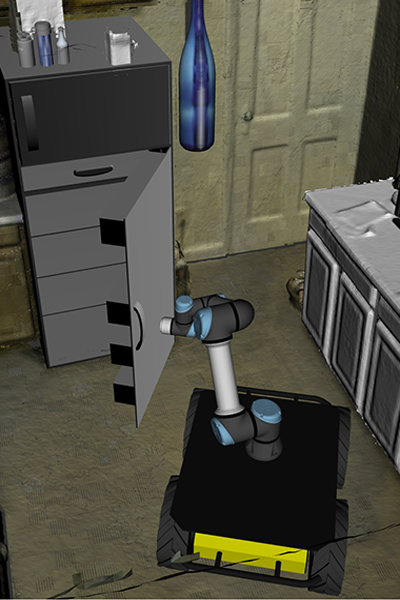}}
            }
        \end{overpic}
    \end{subfigure}%
    \begin{subfigure}[c]{0.2\linewidth}
        \begin{overpic}[width=\linewidth]{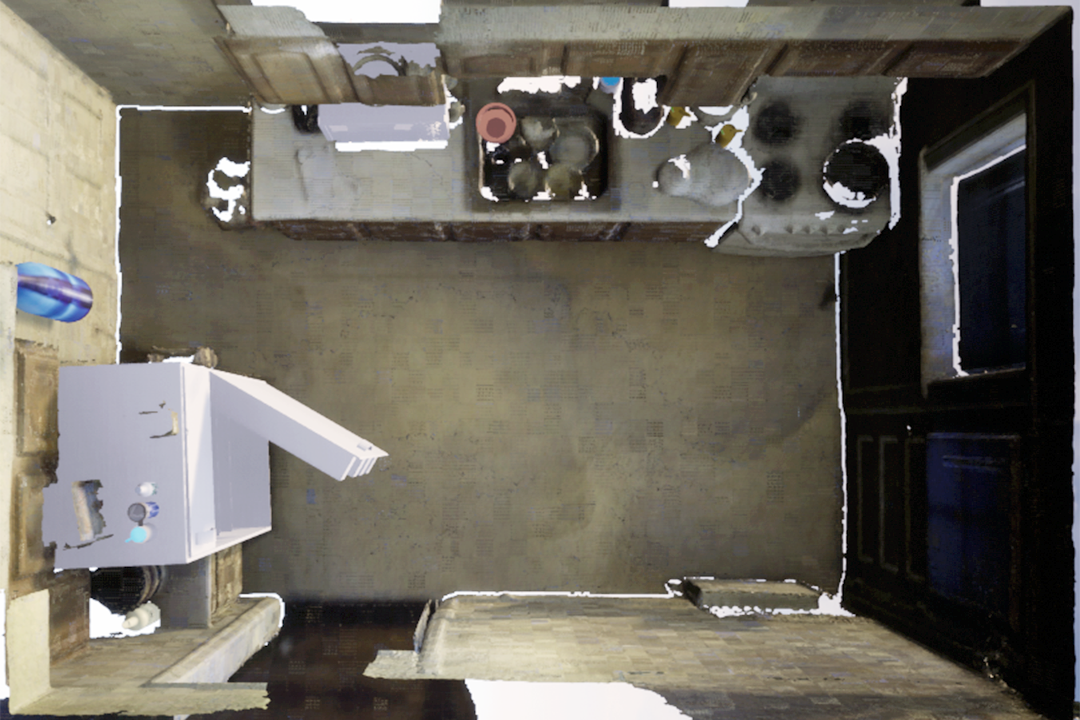}
            \put(59,1){\color{white}\linethickness{0.5mm}%
                \frame{\includegraphics[width=0.4\linewidth]{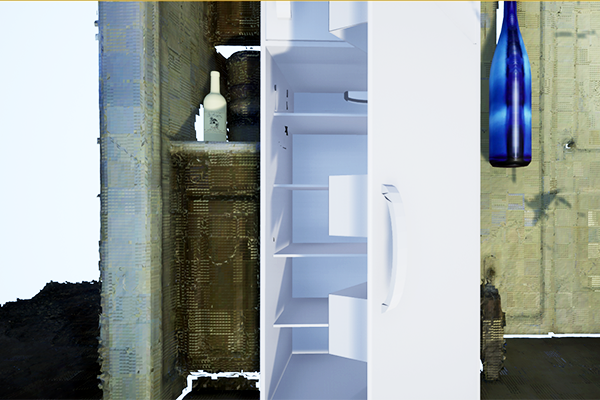}}
            }
        \end{overpic}
    \end{subfigure}%
    \\
    \begin{subfigure}[c]{0.2\linewidth}
        \includegraphics[width=\linewidth]{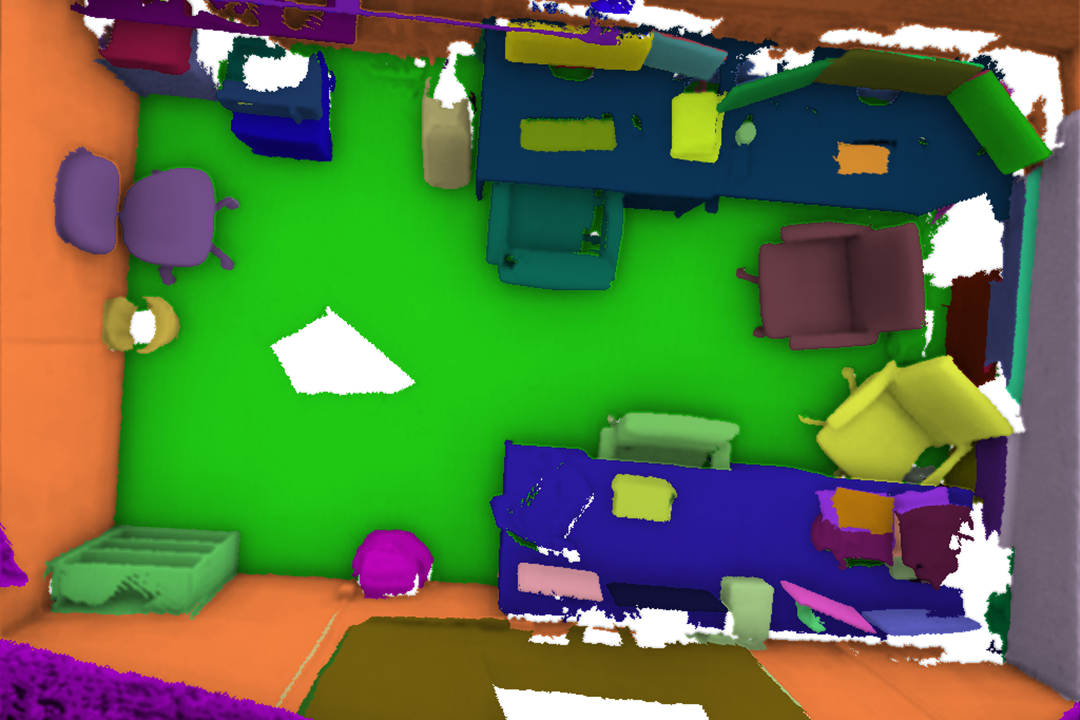}
    \end{subfigure}%
    \begin{subfigure}[c]{0.2\linewidth}
        \includegraphics[width=\linewidth]{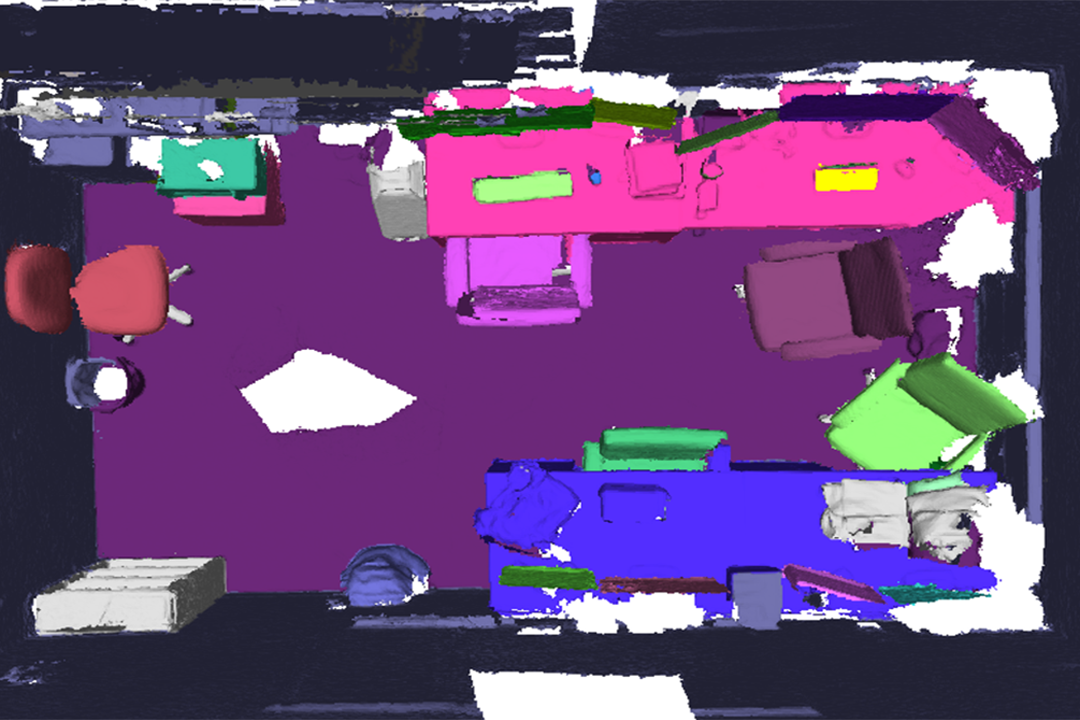}
    \end{subfigure}%
    \begin{subfigure}[c]{0.2\linewidth}
        \includegraphics[width=\linewidth]{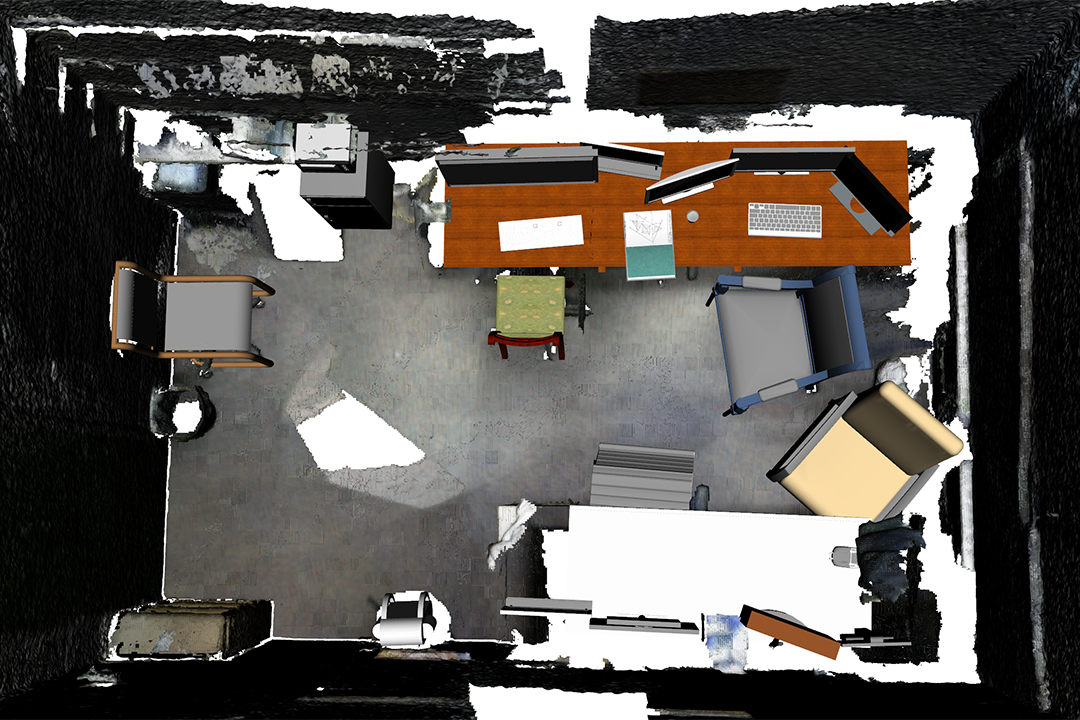}
    \end{subfigure}%
    \begin{subfigure}[c]{0.2\linewidth}
        \begin{overpic}[width=\linewidth]{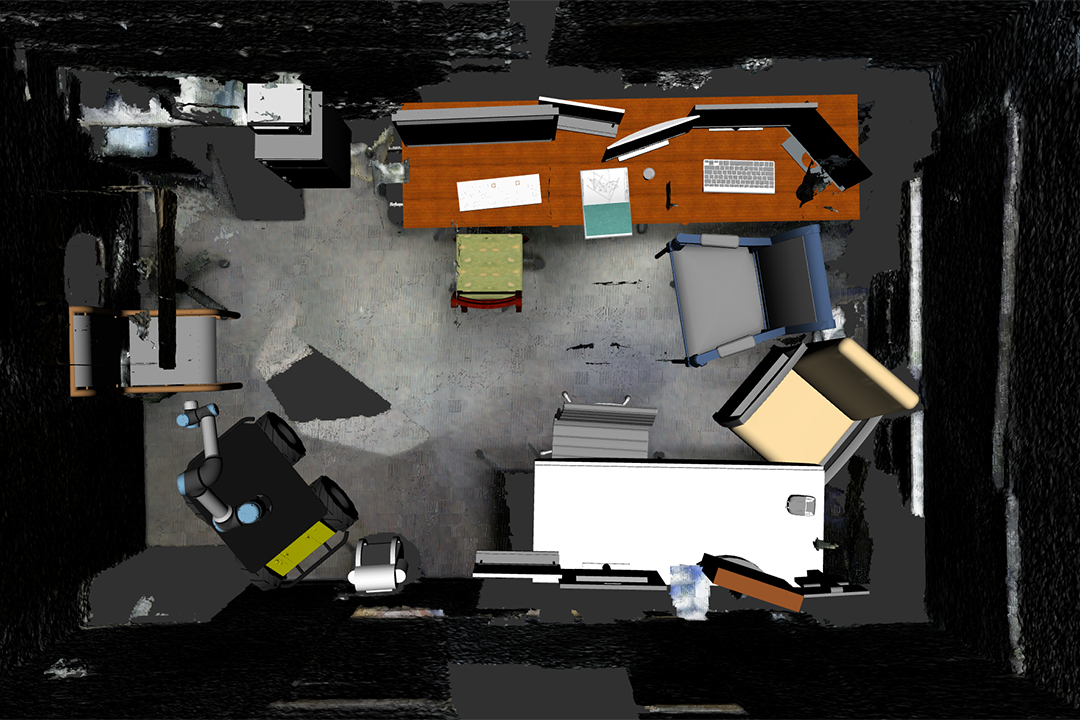}
            \put(68,1){\color{white}\linethickness{0.5mm}%
                \frame{\includegraphics[width=0.3\linewidth]{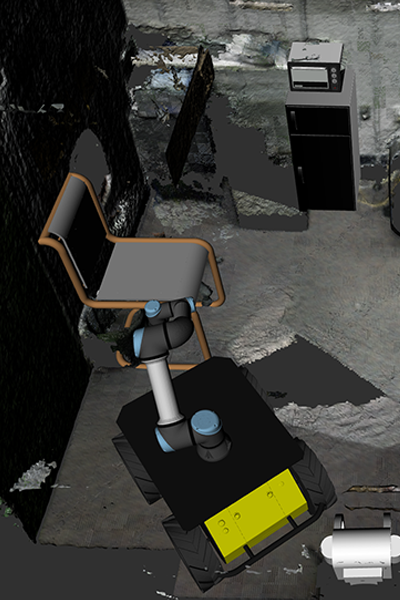}}
            }
        \end{overpic}
    \end{subfigure}%
    \begin{subfigure}[c]{0.2\linewidth}
        \begin{overpic}[width=\linewidth]{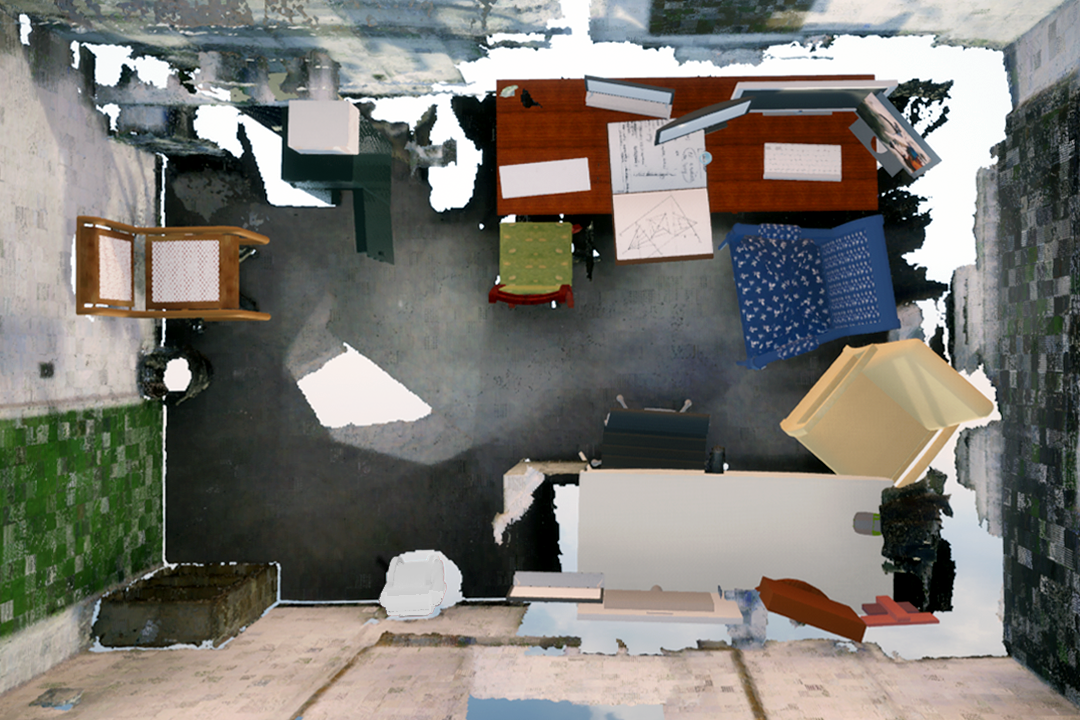}
            \put(59,1){\color{white}\linethickness{0.5mm}%
                \frame{\includegraphics[width=0.4\linewidth]{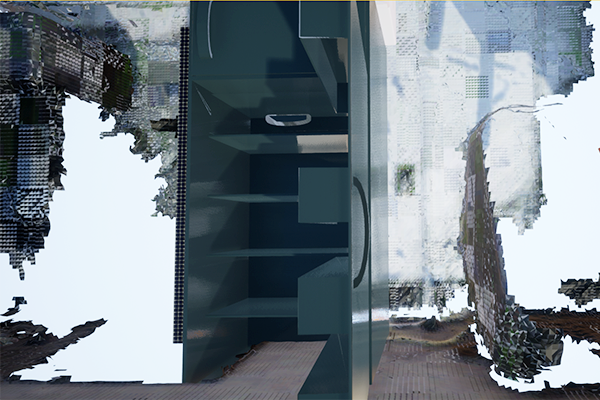}}
            }
        \end{overpic}
    \end{subfigure}%
    \\
    \begin{subfigure}[c]{0.2\linewidth}
        \includegraphics[width=\linewidth]{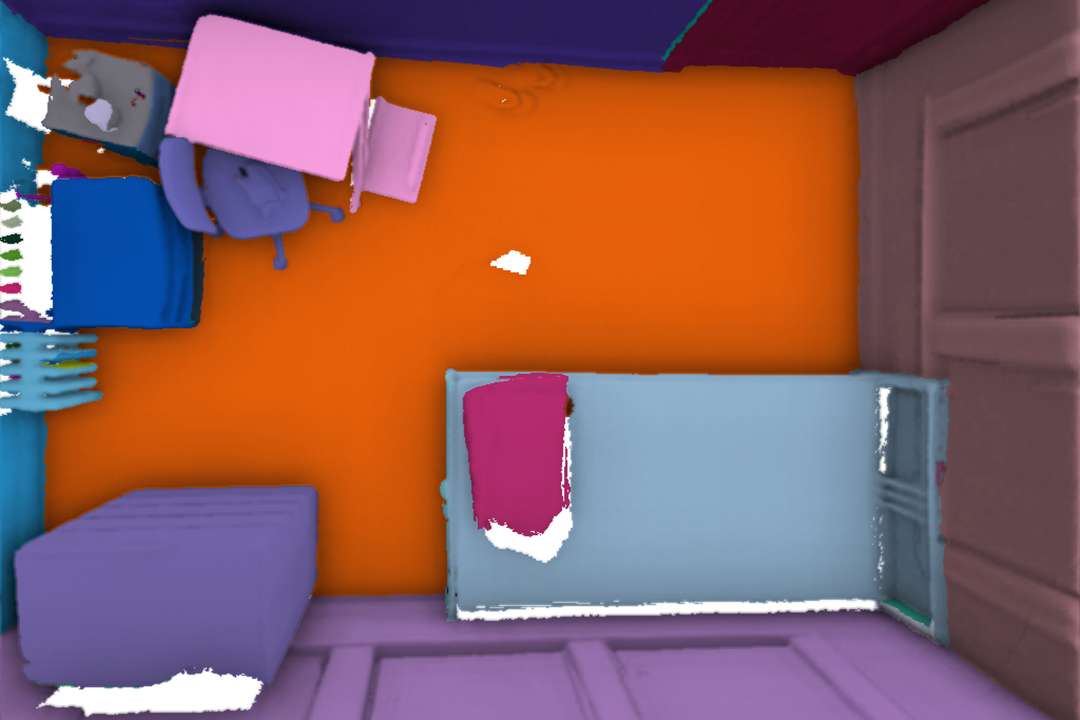}
    \end{subfigure}%
    \begin{subfigure}[c]{0.2\linewidth}
        \includegraphics[width=\linewidth]{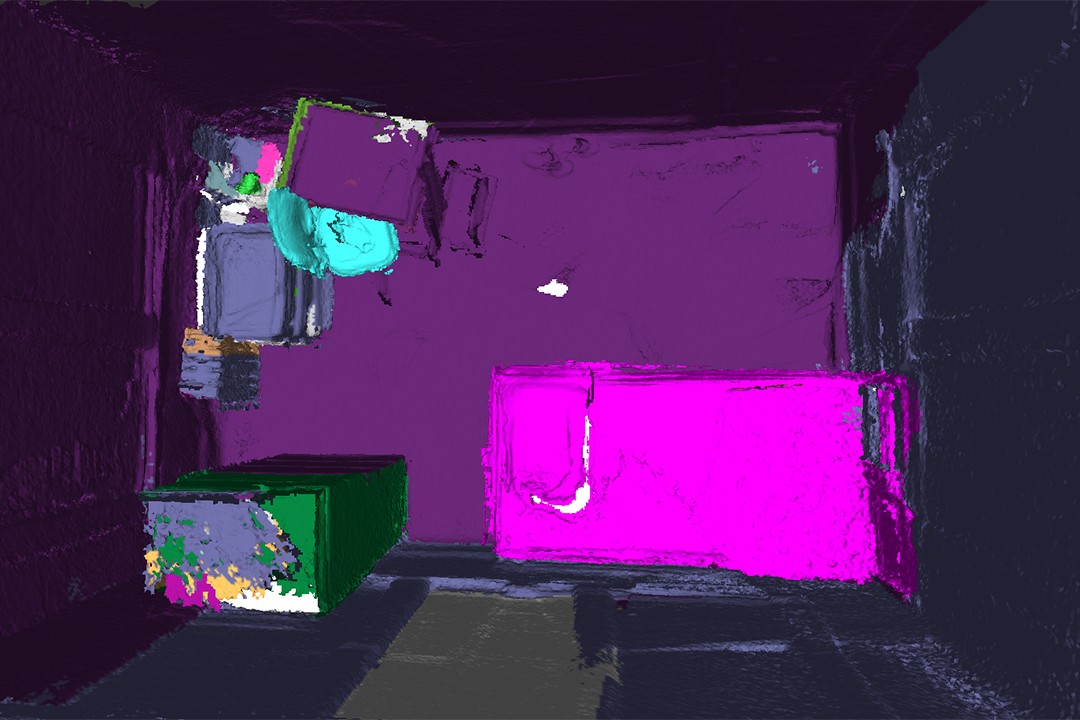}
    \end{subfigure}%
    \begin{subfigure}[c]{0.2\linewidth}
        \includegraphics[width=\linewidth]{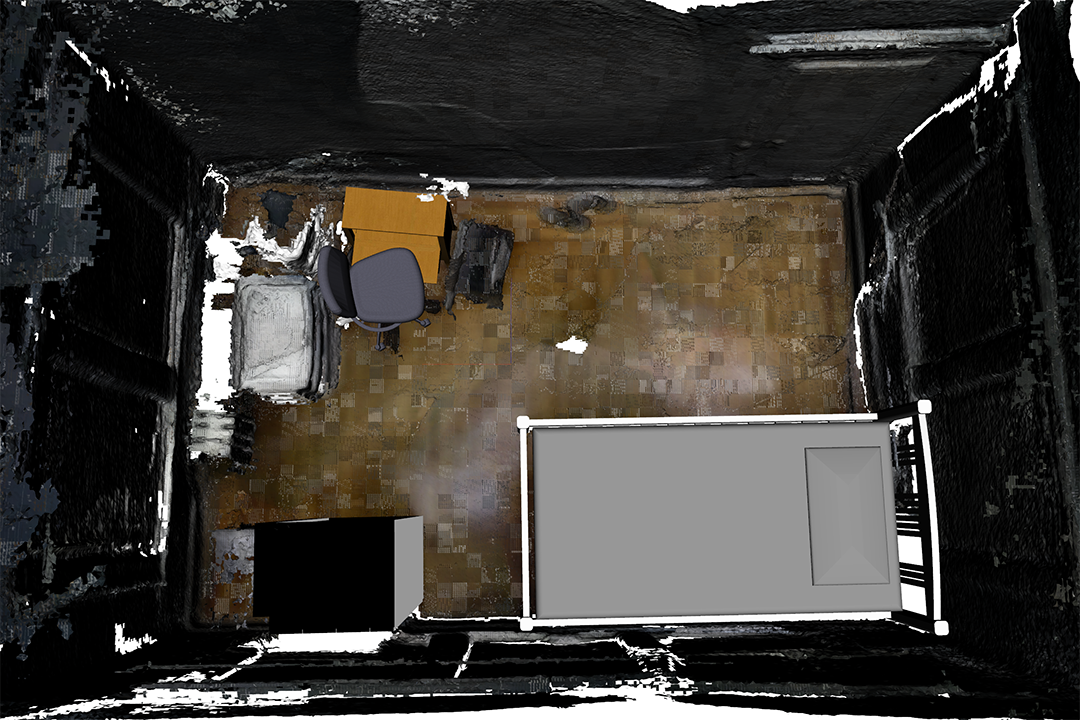}
    \end{subfigure}%
    \begin{subfigure}[c]{0.2\linewidth}
        \begin{overpic}[width=\linewidth]{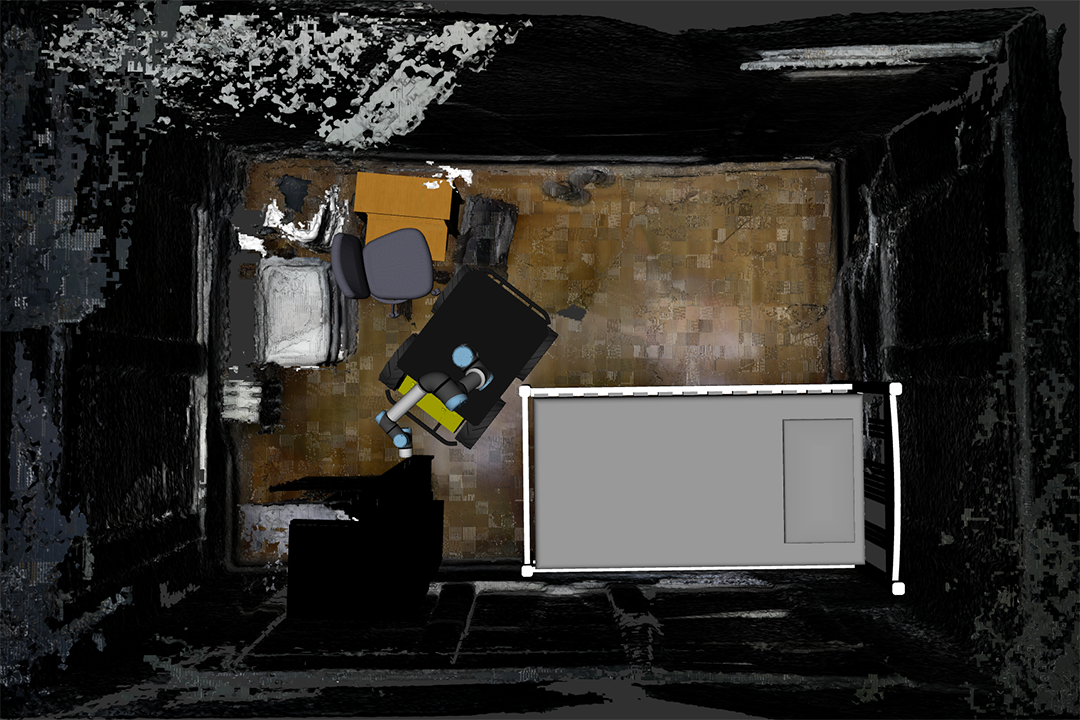}
            \put(68,21){\color{white}\linethickness{0.5mm}%
                \frame{\includegraphics[width=0.3\linewidth]{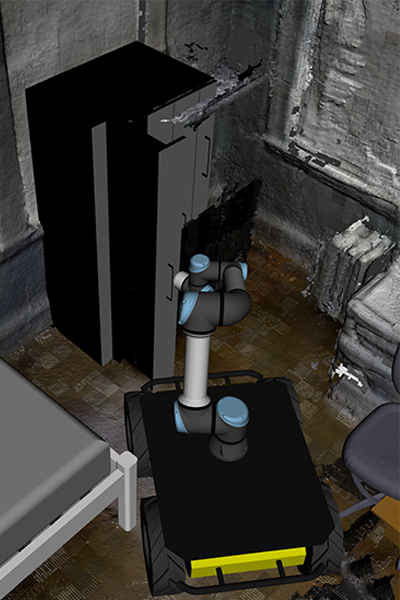}}
            }
        \end{overpic}
    \end{subfigure}%
    \begin{subfigure}[c]{0.2\linewidth}
        \begin{overpic}[width=\linewidth]{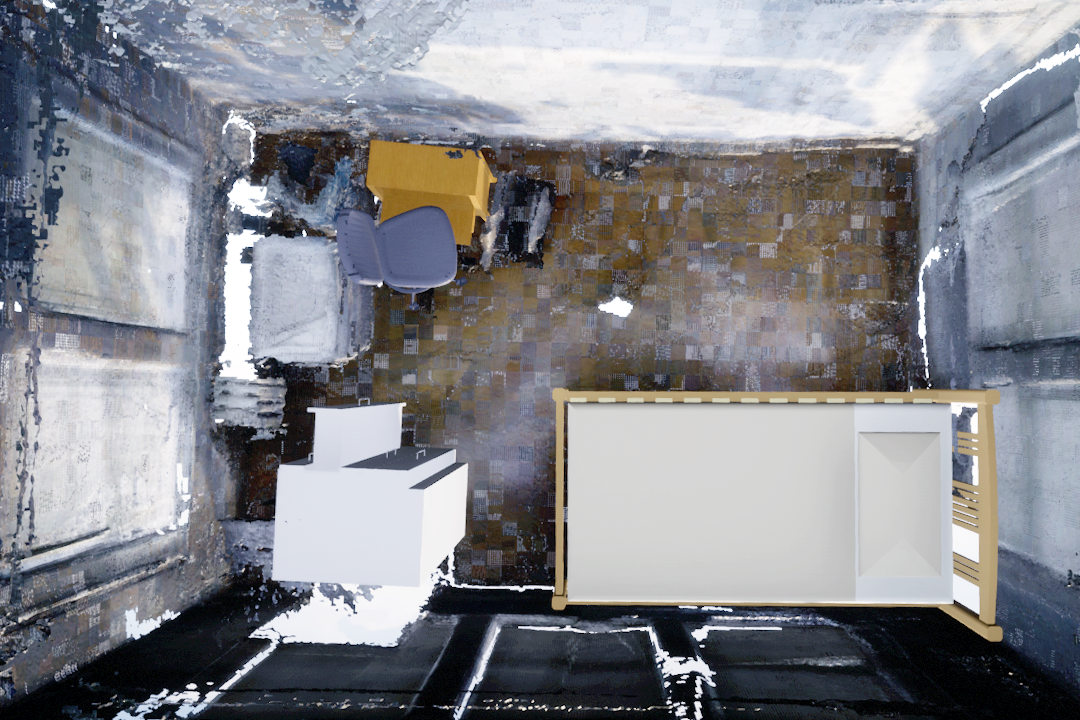}
            \put(59,39){\color{white}\linethickness{0.5mm}%
                \frame{\includegraphics[width=0.4\linewidth]{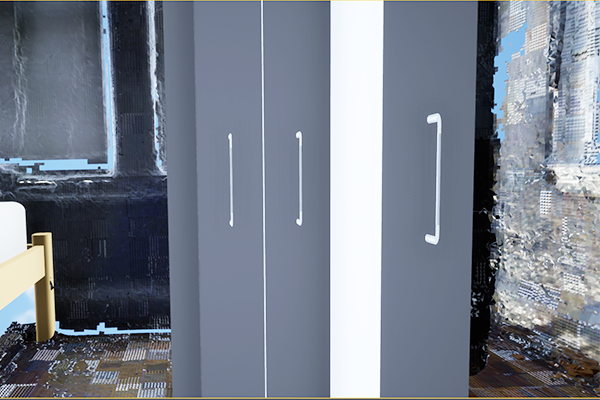}}
            }
        \end{overpic}
    \end{subfigure}%
    \\
    \begin{subfigure}[c]{0.2\linewidth}
        \includegraphics[width=\linewidth]{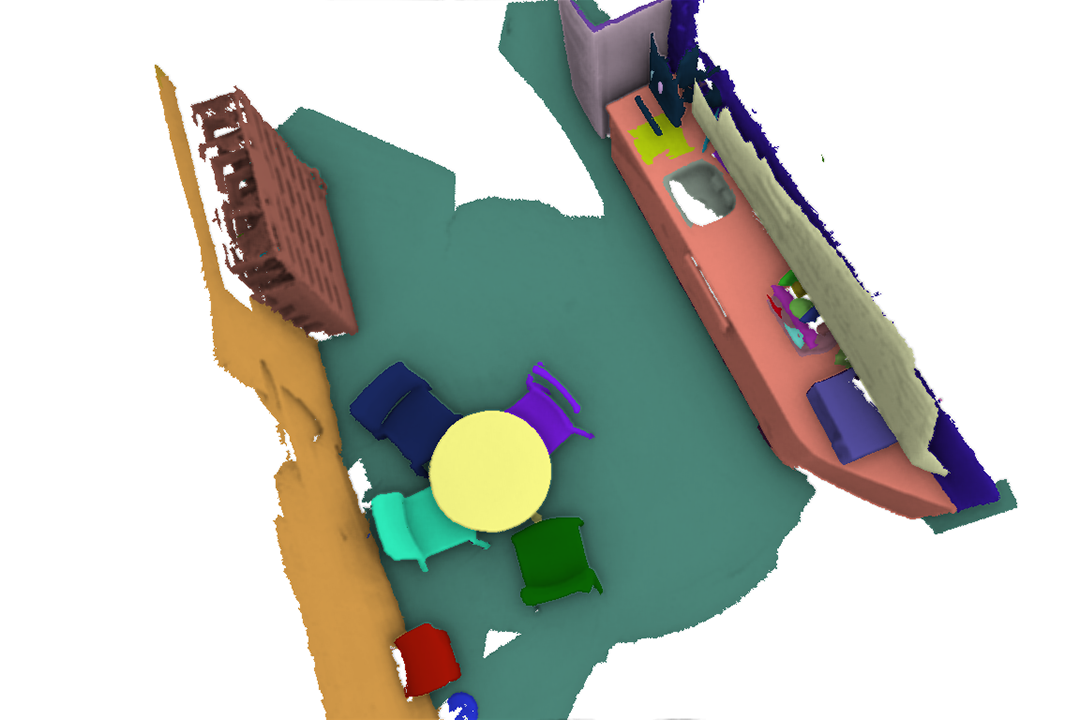}
        \caption{ground-truth segmentation}
    \end{subfigure}%
    \begin{subfigure}[c]{0.2\linewidth}
        \includegraphics[width=\linewidth]{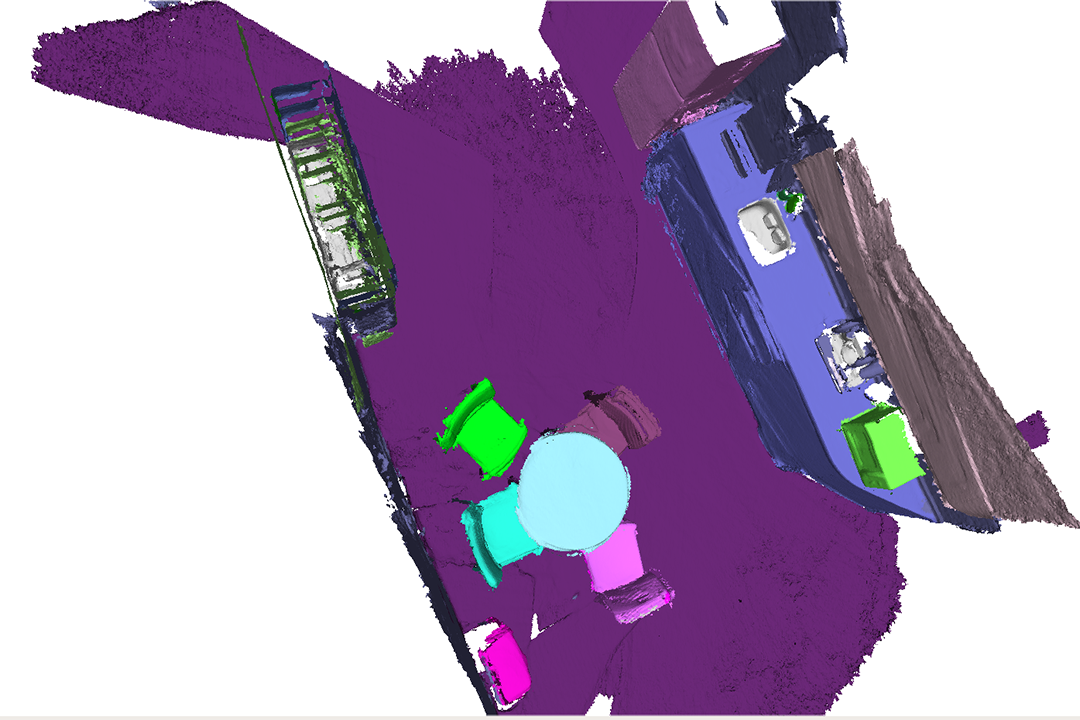}
        \caption{panoptic mapping}
    \end{subfigure}%
    \begin{subfigure}[c]{0.2\linewidth}
        \includegraphics[width=\linewidth]{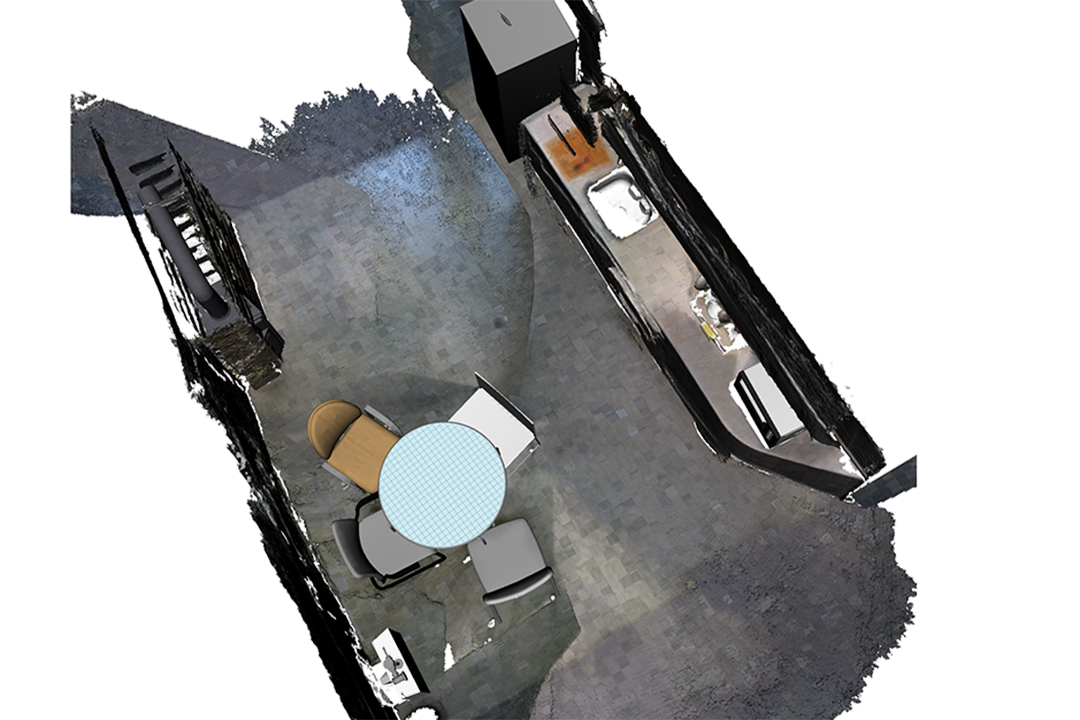}
        \caption{interactive scenes}
    \end{subfigure}%
    \begin{subfigure}[c]{0.2\linewidth}
        \begin{overpic}[width=\linewidth]{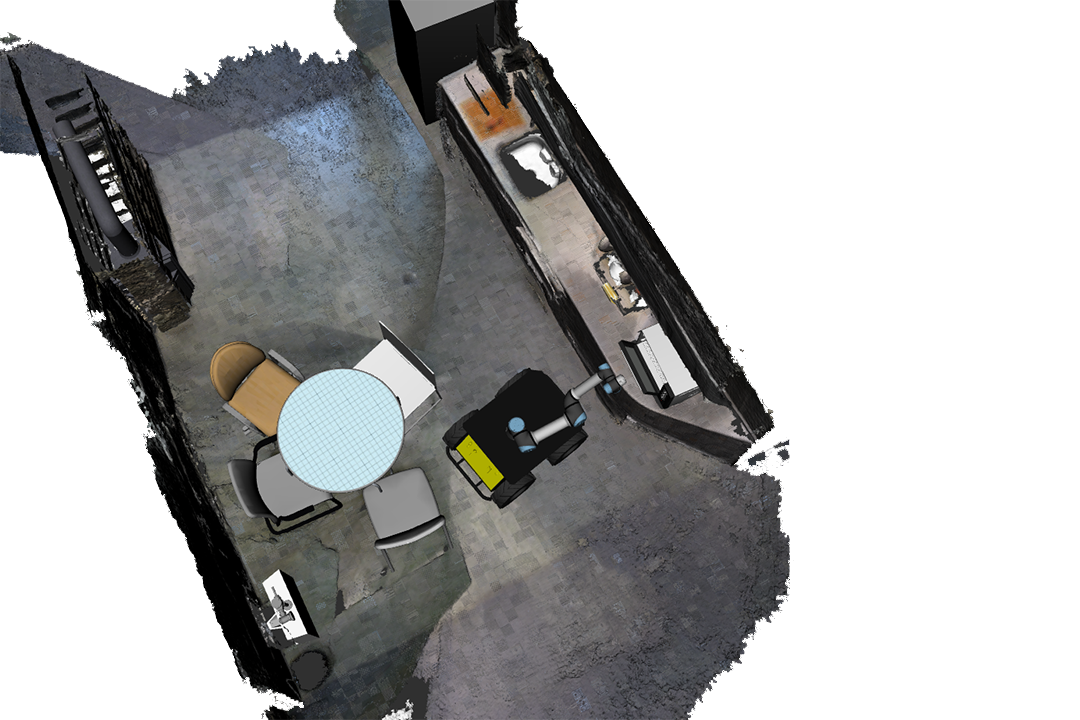}
            \put(68,21){\color{white}\linethickness{0.5mm}%
                \frame{\includegraphics[width=0.3\linewidth]{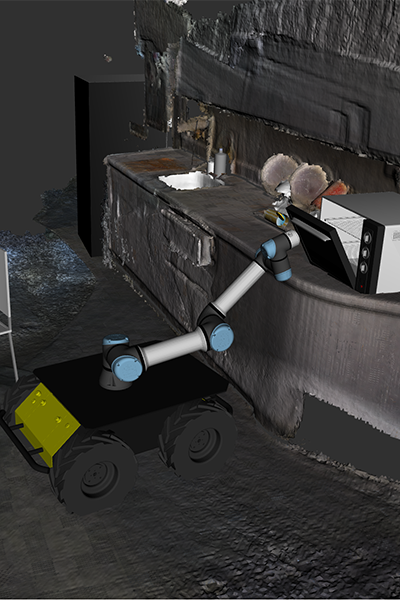}}
            }
        \end{overpic}
        \caption{robot interaction}
    \end{subfigure}%
    \begin{subfigure}[c]{0.2\linewidth}
        \begin{overpic}[width=\linewidth]{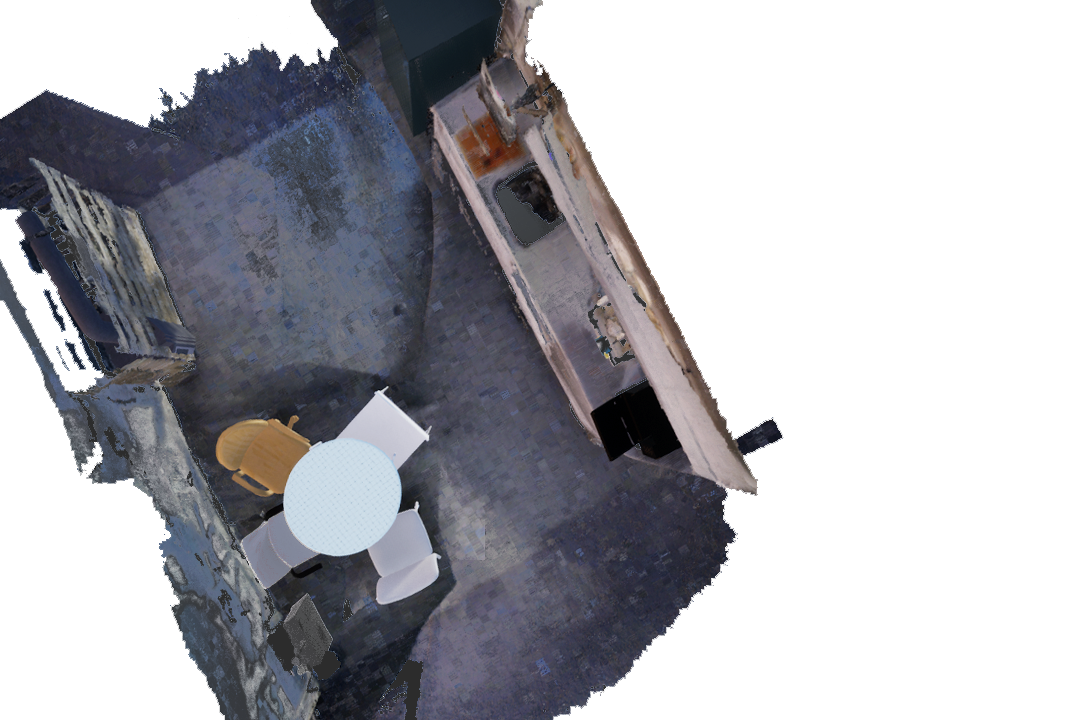}
            \put(59,39){\color{white}\linethickness{0.5mm}%
                \frame{\includegraphics[width=0.4\linewidth]{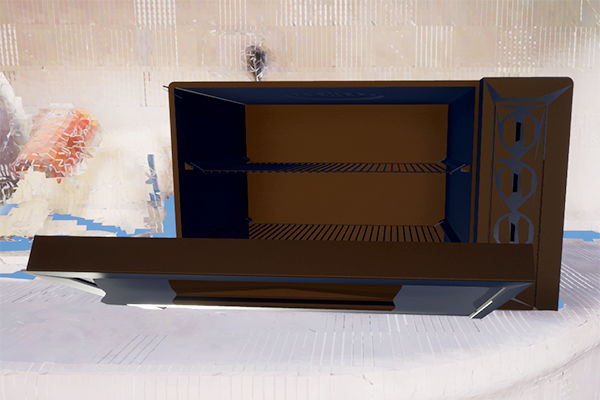}}
            }
        \end{overpic}
        \caption{VR interaction}
    \end{subfigure}%
    \caption{(a--b) Qualitative comparisons between the ground-truth segmentation~\cite{hua2016scenenn} and segmentation results produced by the proposed panoptic mapping. (c) The reconstructed functionally equivalent scenes capture most of the objects and replaces them by actionable CAD models. (d--e) Both robots and human users can virtually enter the reconstructed scene for \ac{tamp} and VR applications, respectively.}
    \label{fig:scene_results}
\end{figure*}

\subsection{Global Physical Violation Check}

Given a shortlist of matched and aligned CAD candidates, we validate supporting relations and proximal relations; see \cref{fig:alignment} for qualitative results. Specifically, for an object node $v_p$ and its object entity $x$, we discard a CAD candidate $y$ if it fails to satisfy \cref{equ:supportingarea} with any supporting child $v_c$ of $v_p$. We also check the proximal constraint by first discarding CAD candidates that collide with the layout entities, and then jointly selecting CAD candidates for each object entity to guarantee the object-object non-collision. The joint selection problem can be formulated as a constraint satisfaction problem. Starting with a CAD candidate with the minimum alignment error for each object entity, we adopt the min-conflict algorithm~\cite{minton1992minimizing} to obtain a global solution.

\section{Experiments and Results}\label{sec:result}




We perform scene reconstruction experiments using RGB-D sequences in the SceneNN dataset~\cite{hua2016scenenn} and import the results into various simulators for interaction; see \cref{fig:scene_results}. Compared to the ground-truth segmentation, our panoptic mapping system accurately recognizes and segments scene entities (\cref{fig:scene_results}b). Such an accurate mapping provides the basis for high-level physical reasoning to replace incomplete meshes with CAD models, resulting in a high-quality, functionally equivalent, interactive scene reconstruction, as shown in \cref{fig:scene_results}c. Note that our system's performance could be further improved as we only utilize pre-trained models in the mapping procedure without fine-tuning. The run-time for converting a 3D panoptic map into an interactive scene varies from 30 seconds to several minutes, depending on the number and categories of functional objects involved.

The reconstructed scene $cg$ can be readily converted into a \ac{urdf} and be imported into robot simulators. While it is straightforward to immigrate scene entities in $cg$ to links and joints in the kinematic tree, supporting edges are altered to fixed/floating joints based on the semantics of the scene entity pairs (\eg, a cup is connected to a table using a floating joint as it can be freely manipulated). \cref{fig:scene_results}c shows the reconstructed scenes in the ROS environment, which subsequently connects the reconstructed scenes and robot \ac{tamp}; see \cref{fig:scene_results}d. \cref{fig:scene_results}e demonstrates that the reconstructed scenes can be loaded into the VR environment~\cite{xie2019vrgym} for interactions with both virtual agents and human users, which opens a new avenue for future studies.

\section{Conclusions}\label{sec:conclusion}

We proposed a new task of reconstructing interactive scenes that captures the semantic and associated actionable information of objects in a scene, instead of purely focusing on geometric reconstruction accuracy. We solved this new task by combining (i) a novel robust panoptic mapping that segments individual objects and layouts, and (ii) a physical reasoning process to replace incomplete objects meshes with part-based CAD models, resulting in physically plausible and interactive scenes. We validated the capability of our system with both qualitative and quantitative results. Finally, we showed that various simulators (\eg, ROS, VR environments) could seamlessly import the reconstructed scene to facilitate researches in robot \ac{tamp} and embodied AI.

This work also motivates three new research questions worth investigating in the future: (i) To sufficiently plan robot tasks, how well should the CAD models replicate the physical objects? (ii) Although the proposed system can filter out dynamic entities based on their semantic segmentation (\eg, humans) and a better data association can handle semi-dynamic objects, how could we incorporate the causal relations between environmental changes and human activities? (iii) Although the effects of acting in a sequential task could be updated as the kinematic information in $cg$, recognizing these effects in physical world introduces extra challenges.

\setstretch{1}
\balance
\bibliographystyle{ieeetr}
\bibliography{IEEEfull}

\begin{thebibliography}{10}

\bibitem{gibson1950perception}
J.~J. Gibson, {\em The perception of the visual world.}
\newblock Houghton Mifflin, 1950.

\bibitem{gibson1966senses}
J.~J. Gibson, {\em The senses considered as perceptual systems.}
\newblock Houghton Mifflin, 1966.

\bibitem{ikeuchi1992task}
K.~Ikeuchi and M.~Hebert, ``Task oriented vision,'' in {\em Proceedings of
  International Conference on Intelligent Robots and Systems (IROS)}, 1992.

\bibitem{zhu2015understanding}
Y.~Zhu, Y.~Zhao, and S.-C. Zhu, ``Understanding tools: Task-oriented object
  modeling, learning and recognition,'' in {\em Proceedings of the IEEE
  Conference on Computer Vision and Pattern Recognition (CVPR)}, 2015.

\bibitem{hoang2020panoptic}
D.-C. Hoang, A.~J. Lilienthal, and T.~Stoyanov, ``Panoptic 3d mapping and
  object pose estimation using adaptively weighted semantic information,'' {\em
  Robotics and Automation Letters (RA-L)}, vol.~5, no.~2, pp.~1962--1969, 2020.

\bibitem{narita2019panopticfusion}
G.~Narita, T.~Seno, T.~Ishikawa, and Y.~Kaji, ``Panopticfusion: Online
  volumetric semantic mapping at the level of stuff and things,'' in {\em
  Proceedings of International Conference on Intelligent Robots and Systems
  (IROS)}, 2019.

\bibitem{myers2015affordance}
A.~Myers, C.~L. Teo, C.~Ferm{\"u}ller, and Y.~Aloimonos, ``Affordance detection
  of tool parts from geometric features,'' in {\em Proceedings of International
  Conference on Robotics and Automation (ICRA)}, 2015.

\bibitem{zhu2016inferring}
Y.~Zhu, C.~Jiang, Y.~Zhao, D.~Terzopoulos, and S.-C. Zhu, ``Inferring forces
  and learning human utilities from videos,'' in {\em Proceedings of the IEEE
  Conference on Computer Vision and Pattern Recognition (CVPR)}, 2016.

\bibitem{zheng2013beyond}
B.~Zheng, Y.~Zhao, J.~C. Yu, K.~Ikeuchi, and S.-C. Zhu, ``Beyond point clouds:
  Scene understanding by reasoning geometry and physics,'' in {\em Proceedings
  of the IEEE Conference on Computer Vision and Pattern Recognition (CVPR)},
  2013.

\bibitem{zheng2014detecting}
B.~Zheng, Y.~Zhao, C.~Y. Joey, K.~Ikeuchi, and S.-C. Zhu, ``Detecting potential
  falling objects by inferring human action and natural disturbance,'' in {\em
  Proceedings of International Conference on Robotics and Automation (ICRA)},
  2014.

\bibitem{zhao2013scene}
Y.~Zhao and S.-C. Zhu, ``Scene parsing by integrating function, geometry and
  appearance models,'' in {\em Proceedings of the IEEE Conference on Computer
  Vision and Pattern Recognition (CVPR)}, 2013.

\bibitem{zheng2015scene}
B.~Zheng, Y.~Zhao, J.~Yu, K.~Ikeuchi, and S.-C. Zhu, ``Scene understanding by
  reasoning stability and safety,'' {\em International Journal of Robotics
  Research (IJRR)}, vol.~112, no.~2, pp.~221--238, 2015.

\bibitem{huang2018cooperative}
S.~Huang, S.~Qi, Y.~Xiao, Y.~Zhu, Y.~N. Wu, and S.-C. Zhu, ``Cooperative
  holistic scene understanding: Unifying 3d object, layout, and camera pose
  estimation,'' in {\em Proceedings of Advances in Neural Information
  Processing Systems (NeurIPS)}, 2018.

\bibitem{chen2019holistic++}
Y.~Chen, S.~Huang, T.~Yuan, S.~Qi, Y.~Zhu, and S.-C. Zhu, ``Holistic++ scene
  understanding: Single-view 3d holistic scene parsing and human pose
  estimation with human-object interaction and physical commonsense,'' in {\em
  Proceedings of International Conference on Computer Vision (ICCV)}, 2019.

\bibitem{kaelbling2011hierarchical}
L.~P. Kaelbling and T.~Lozano-P{\'e}rez, ``Hierarchical task and motion
  planning in the now,'' in {\em Proceedings of International Conference on
  Robotics and Automation (ICRA)}, 2011.

\bibitem{srivastava2014combined}
S.~Srivastava, E.~Fang, L.~Riano, R.~Chitnis, S.~Russell, and P.~Abbeel,
  ``Combined task and motion planning through an extensible planner-independent
  interface layer,'' in {\em Proceedings of International Conference on
  Robotics and Automation (ICRA)}, 2014.

\bibitem{kim2019learning}
B.~Kim, Z.~Wang, L.~P. Kaelbling, and T.~Lozano-P{\'e}rez, ``Learning to guide
  task and motion planning using score-space representation,'' {\em
  International Journal of Robotics Research (IJRR)}, vol.~38, no.~7,
  pp.~793--812, 2019.

\bibitem{wang2018active}
Z.~Wang, C.~R. Garrett, L.~P. Kaelbling, and T.~Lozano-P{\'e}rez, ``Active
  model learning and diverse action sampling for task and motion planning,'' in
  {\em Proceedings of International Conference on Intelligent Robots and
  Systems (IROS)}, 2018.

\bibitem{xia2020interactive}
F.~Xia, W.~B. Shen, C.~Li, P.~Kasimbeg, M.~E. Tchapmi, A.~Toshev,
  R.~Mart{\'\i}n-Mart{\'\i}n, and S.~Savarese, ``Interactive gibson benchmark:
  A benchmark for interactive navigation in cluttered environments,'' {\em
  Robotics and Automation Letters (RA-L)}, vol.~5, no.~2, pp.~713--720, 2020.

\bibitem{xiang2020sapien}
F.~Xiang, Y.~Qin, K.~Mo, Y.~Xia, H.~Zhu, F.~Liu, M.~Liu, H.~Jiang, Y.~Yuan,
  H.~Wang, {\em et~al.}, ``Sapien: A simulated part-based interactive
  environment,'' in {\em Proceedings of the IEEE Conference on Computer Vision
  and Pattern Recognition (CVPR)}, 2020.

\bibitem{pronobis2012large}
A.~Pronobis and P.~Jensfelt, ``Large-scale semantic mapping and reasoning with
  heterogeneous modalities,'' in {\em Proceedings of International Conference
  on Robotics and Automation (ICRA)}, 2012.

\bibitem{yang2019cubeslam}
S.~Yang and S.~Scherer, ``Cubeslam: Monocular 3-d object slam,'' {\em
  Transactions on Robotics (T-RO)}, vol.~35, no.~4, pp.~925--938, 2019.

\bibitem{mccormac2017semanticfusion}
J.~McCormac, A.~Handa, A.~Davison, and S.~Leutenegger, ``Semanticfusion: Dense
  3d semantic mapping with convolutional neural networks,'' in {\em Proceedings
  of International Conference on Robotics and Automation (ICRA)}, 2017.

\bibitem{grinvald2019volumetric}
M.~Grinvald, F.~Furrer, T.~Novkovic, J.~J. Chung, C.~Cadena, R.~Siegwart, and
  J.~Nieto, ``Volumetric instance-aware semantic mapping and 3d object
  discovery,'' {\em Robotics and Automation Letters (RA-L)}, vol.~4, no.~3,
  pp.~3037--3044, 2019.

\bibitem{mccormac2018fusion++}
J.~McCormac, R.~Clark, M.~Bloesch, A.~Davison, and S.~Leutenegger, ``Fusion++:
  Volumetric object-level slam,'' in {\em Proceedings of International
  Conference on 3D Vision (3DV)}, 2018.

\bibitem{kirillov2019panoptic}
A.~Kirillov, K.~He, R.~Girshick, C.~Rother, and P.~Doll{\'a}r, ``Panoptic
  segmentation,'' in {\em Proceedings of the IEEE Conference on Computer Vision
  and Pattern Recognition (CVPR)}, 2019.

\bibitem{koenig2004design}
N.~P. Koenig and A.~Howard, ``Design and use paradigms for gazebo, an
  open-source multi-robot simulator.,'' in {\em Proceedings of International
  Conference on Intelligent Robots and Systems (IROS)}, 2004.

\bibitem{rohmer2013v}
E.~Rohmer, S.~P. Singh, and M.~Freese, ``V-rep: A versatile and scalable robot
  simulation framework,'' in {\em Proceedings of International Conference on
  Intelligent Robots and Systems (IROS)}, 2013.

\bibitem{kolve2017ai2}
E.~Kolve, R.~Mottaghi, D.~Gordon, Y.~Zhu, A.~Gupta, and A.~Farhadi, ``Ai2-thor:
  An interactive 3d environment for visual ai,'' 2017.

\bibitem{xia2018gibson}
F.~Xia, A.~R. Zamir, Z.~He, A.~Sax, J.~Malik, and S.~Savarese, ``Gibson env:
  Real-world perception for embodied agents,'' in {\em Proceedings of the IEEE
  Conference on Computer Vision and Pattern Recognition (CVPR)}, 2018.

\bibitem{xie2019vrgym}
X.~Xie, H.~Liu, Z.~Zhang, Y.~Qiu, F.~Gao, S.~Qi, Y.~Zhu, and S.-C. Zhu,
  ``Vrgym: A virtual testbed for physical and interactive ai,'' in {\em
  Proceedings of the ACM Turing Celebration Conference-China}, pp.~1--6, 2019.

\bibitem{yu2011make}
L.~F. Yu, S.~K. Yeung, C.~K. Tang, D.~Terzopoulos, T.~F. Chan, and S.~J. Osher,
  ``Make it home: automatic optimization of furniture arrangement,'' {\em ACM
  Transactions on Graphics (TOG)}, vol.~30, no.~4, 2011.

\bibitem{qi2018human}
S.~Qi, Y.~Zhu, S.~Huang, C.~Jiang, and S.-C. Zhu, ``Human-centric indoor scene
  synthesis using stochastic grammar,'' in {\em Proceedings of the IEEE
  Conference on Computer Vision and Pattern Recognition (CVPR)}, 2018.

\bibitem{jiang2018configurable}
C.~Jiang, S.~Qi, Y.~Zhu, S.~Huang, J.~Lin, L.-F. Yu, D.~Terzopoulos, and S.-C.
  Zhu, ``Configurable 3d scene synthesis and 2d image rendering with per-pixel
  ground truth using stochastic grammars,'' {\em International Journal of
  Computer Vision (IJCV)}, vol.~126, no.~9, pp.~920--941, 2018.

\bibitem{song2017semantic}
S.~Song, F.~Yu, A.~Zeng, A.~X. Chang, M.~Savva, and T.~Funkhouser, ``Semantic
  scene completion from a single depth image,'' in {\em Proceedings of the IEEE
  Conference on Computer Vision and Pattern Recognition (CVPR)}, 2017.

\bibitem{fu20203dfront}
H.~Fu, B.~Cai, L.~Gao, L.~Zhang, C.~Li, Q.~Zeng, C.~Sun, Y.~Fei, Y.~Zheng,
  Y.~Li, Y.~Liu, P.~Liu, L.~Ma, L.~Weng, X.~Hu, X.~Ma, Q.~Qian, R.~Jia,
  B.~Zhao, and H.~Zhang, ``3d-front: 3d furnished rooms with layouts and
  semantics,'' {\em arXiv preprint arXiv:2011.09127}, 2020.

\bibitem{pham2019real}
Q.-H. Pham, B.-S. Hua, T.~Nguyen, and S.-K. Yeung, ``Real-time progressive 3d
  semantic segmentation for indoor scenes,'' in {\em Winter Conference on
  Applications of Computer Vision (WACV)}, 2019.

\bibitem{yang2019monocular}
S.~Yang and S.~Scherer, ``Monocular object and plane slam in structured
  environments,'' {\em Robotics and Automation Letters (RA-L)}, vol.~4, no.~4,
  pp.~3145--3152, 2019.

\bibitem{wada2020morefusion}
K.~Wada, E.~Sucar, S.~James, D.~Lenton, and A.~J. Davison, ``Morefusion:
  Multi-object reasoning for 6d pose estimation from volumetric fusion,'' in
  {\em Proceedings of the IEEE Conference on Computer Vision and Pattern
  Recognition (CVPR)}, 2020.

\bibitem{sui2020geofusion}
Z.~Sui, H.~Chang, N.~Xu, and O.~Chadwicke~Jenkins, ``Geofusion: Geometric
  consistency informed scene estimation in dense clutter,'' {\em Robotics and
  Automation Letters (RA-L)}, 2020.

\bibitem{yi2019gspn}
L.~Yi, W.~Zhao, H.~Wang, M.~Sung, and L.~J. Guibas, ``Gspn: Generative shape
  proposal network for 3d instance segmentation in point cloud,'' in {\em
  Proceedings of the IEEE Conference on Computer Vision and Pattern Recognition
  (CVPR)}, 2019.

\bibitem{pham2019jsis3d}
Q.-H. Pham, T.~Nguyen, B.-S. Hua, G.~Roig, and S.-K. Yeung, ``Jsis3d: joint
  semantic-instance segmentation of 3d point clouds with multi-task pointwise
  networks and multi-value conditional random fields,'' in {\em Proceedings of
  the IEEE Conference on Computer Vision and Pattern Recognition (CVPR)}, 2019.

\bibitem{dai2017scannet}
A.~Dai, A.~X. Chang, M.~Savva, M.~Halber, T.~Funkhouser, and M.~Nie{\ss}ner,
  ``Scannet: Richly-annotated 3d reconstructions of indoor scenes,'' in {\em
  Proceedings of the IEEE Conference on Computer Vision and Pattern Recognition
  (CVPR)}, 2017.

\bibitem{avetisyan2019scan2cad}
A.~Avetisyan, M.~Dahnert, A.~Dai, M.~Savva, A.~X. Chang, and M.~Nie{\ss}ner,
  ``Scan2cad: Learning cad model alignment in rgb-d scans,'' in {\em
  Proceedings of the IEEE Conference on Computer Vision and Pattern Recognition
  (CVPR)}, 2019.

\bibitem{avetisyan2020scenecad}
A.~Avetisyan, T.~Khanova, C.~Choy, D.~Dash, A.~Dai, and M.~Nießner,
  ``Scenecad: Predicting object alignments and layouts in rgb-d scans,'' in
  {\em The European Conference on Computer Vision (ECCV)}, August 2020.

\bibitem{cadena2016past}
C.~Cadena, L.~Carlone, H.~Carrillo, Y.~Latif, D.~Scaramuzza, J.~Neira, I.~Reid,
  and J.~J. Leonard, ``Past, present, and future of simultaneous localization
  and mapping: Toward the robust-perception age,'' {\em Transactions on
  Robotics (T-RO)}, vol.~32, no.~6, pp.~1309--1332, 2016.

\bibitem{zhu2007stochastic}
S.-C. Zhu and D.~Mumford, ``A stochastic grammar of images,'' {\em Foundations
  and Trends{\textregistered} in Computer Graphics and Vision}, vol.~2, no.~4,
  pp.~259--362, 2007.

\bibitem{zhao2011image}
Y.~Zhao and S.-C. Zhu, ``Image parsing with stochastic scene grammar,'' in {\em
  Proceedings of Advances in Neural Information Processing Systems (NeurIPS)},
  2011.

\bibitem{armeni20193d}
I.~Armeni, Z.-Y. He, J.~Gwak, A.~R. Zamir, M.~Fischer, J.~Malik, and
  S.~Savarese, ``3d scene graph: A structure for unified semantics, 3d space,
  and camera,'' in {\em Proceedings of the IEEE Conference on Computer Vision
  and Pattern Recognition (CVPR)}, 2019.

\bibitem{wald2020learning}
J.~Wald, H.~Dhamo, N.~Navab, and F.~Tombari, ``Learning 3d semantic scene
  graphs from 3d indoor reconstructions,'' in {\em Proceedings of the IEEE
  Conference on Computer Vision and Pattern Recognition (CVPR)}, 2020.

\bibitem{rosinol20203d}
A.~Rosinol, A.~Gupta, M.~Abate, J.~Shi, and L.~Carlone, ``{3D} dynamic scene
  graphs: Actionable spatial perception with places, objects, and humans,'' in
  {\em Proceedings of Robotics: Science and Systems (RSS)}, 2020.

\bibitem{huang2018holistic}
S.~Huang, S.~Qi, Y.~Zhu, Y.~Xiao, Y.~Xu, and S.-C. Zhu, ``Holistic 3d scene
  parsing and reconstruction from a single rgb image,'' in {\em Proceedings of
  European Conference on Computer Vision (ECCV)}, 2018.

\bibitem{qi2020generalized}
S.~Qi, B.~Jia, S.~Huang, P.~Wei, and S.-C. Zhu, ``A generalized earley parser
  for human activity parsing and prediction,'' {\em IEEE Transactions on
  Pattern Analysis and Machine Intelligence (TPAMI)}, 2020.

\bibitem{edmonds2017feeling}
M.~Edmonds, F.~Gao, X.~Xie, H.~Liu, S.~Qi, Y.~Zhu, B.~Rothrock, and S.-C. Zhu,
  ``Feeling the force: Integrating force and pose for fluent discovery through
  imitation learning to open medicine bottles,'' in {\em Proceedings of
  International Conference on Intelligent Robots and Systems (IROS)}, 2017.

\bibitem{liu2018interactive}
H.~Liu, Y.~Zhang, W.~Si, X.~Xie, Y.~Zhu, and S.-C. Zhu, ``Interactive robot
  knowledge patching using augmented reality,'' in {\em Proceedings of
  International Conference on Robotics and Automation (ICRA)}, 2018.

\bibitem{edmonds2019tale}
M.~Edmonds, F.~Gao, H.~Liu, X.~Xie, S.~Qi, B.~Rothrock, Y.~Zhu, Y.~N. Wu,
  H.~Lu, and S.-C. Zhu, ``A tale of two explanations: Enhancing human trust by
  explaining robot behavior,'' {\em Science Robotics}, vol.~4, no.~37, 2019.

\bibitem{liu2019mirroring}
H.~Liu, C.~Zhang, Y.~Zhu, C.~Jiang, and S.-C. Zhu, ``Mirroring without
  overimitation: Learning functionally equivalent manipulation actions,'' in
  {\em Proceedings of AAAI Conference on Artificial Intelligence (AAAI)}, 2019.

\bibitem{zhang2020graph}
Z.~Zhang, Y.~Zhu, and S.-C. Zhu, ``Graph-based hierarchical knowledge
  representation for robot task transfer from virtual to physical world,'' in
  {\em Proceedings of International Conference on Intelligent Robots and
  Systems (IROS)}, 2020.

\bibitem{yuan2020joint}
T.~Yuan, H.~Liu, L.~Fan, Z.~Zheng, T.~Gao, Y.~Zhu, and S.-C. Zhu, ``Joint
  inference of states, robot knowledge, and human (false-) beliefs,'' in {\em
  Proceedings of International Conference on Intelligent Robots and Systems
  (IROS)}, 2020.

\bibitem{hartley2003multiple}
R.~Hartley and A.~Zisserman, {\em Multiple view geometry in computer vision}.
\newblock Cambridge university press, 2003.

\bibitem{taguchi2013point}
Y.~Taguchi, Y.-D. Jian, S.~Ramalingam, and C.~Feng, ``Point-plane slam for
  hand-held 3d sensors,'' in {\em Proceedings of International Conference on
  Robotics and Automation (ICRA)}, 2013.

\bibitem{malandain2002computing}
G.~Malandain and J.-D. Boissonnat, ``Computing the diameter of a point set,''
  {\em International Journal of Computational Geometry \& Applications},
  vol.~12, no.~06, pp.~489--509, 2002.

\bibitem{wu2019detectron2}
Y.~Wu, A.~Kirillov, F.~Massa, W.-Y. Lo, and R.~Girshick, ``Detectron2.''
  \url{https://github.com/facebookresearch/detectron2}, 2019.

\bibitem{lin2014microsoft}
T.-Y. Lin, M.~Maire, S.~Belongie, J.~Hays, P.~Perona, D.~Ramanan,
  P.~Doll{\'a}r, and C.~L. Zitnick, ``Microsoft coco: Common objects in
  context,'' in {\em Proceedings of European Conference on Computer Vision
  (ECCV)}, 2014.

\bibitem{furrer2018incremental}
F.~Furrer, T.~Novkovic, M.~Fehr, A.~Gawel, M.~Grinvald, T.~Sattler,
  R.~Siegwart, and J.~Nieto, ``Incremental object database: Building 3d models
  from multiple partial observations,'' in {\em Proceedings of International
  Conference on Intelligent Robots and Systems (IROS)}, 2018.

\bibitem{hua2016scenenn}
B.-S. Hua, Q.-H. Pham, D.~T. Nguyen, M.-K. Tran, L.-F. Yu, and S.-K. Yeung,
  ``Scenenn: A scene meshes dataset with annotations,'' in {\em Proceedings of
  International Conference on 3D Vision (3DV)}, 2016.

\bibitem{chang2015shapenet}
A.~X. Chang, T.~Funkhouser, L.~Guibas, P.~Hanrahan, Q.~Huang, Z.~Li,
  S.~Savarese, M.~Savva, S.~Song, H.~Su, {\em et~al.}, ``Shapenet: An
  information-rich 3d model repository,'' {\em arXiv preprint
  arXiv:1512.03012}, 2015.

\bibitem{jonker1987shortest}
R.~Jonker and A.~Volgenant, ``A shortest augmenting path algorithm for dense
  and sparse linear assignment problems,'' {\em Computing}, vol.~38, no.~4,
  pp.~325--340, 1987.

\bibitem{more1978levenberg}
J.~J. Mor{\'e}, ``The levenberg-marquardt algorithm: implementation and
  theory,'' in {\em Numerical analysis}, pp.~105--116, Springer, 1978.

\bibitem{minton1992minimizing}
S.~Minton, M.~D. Johnston, A.~B. Philips, and P.~Laird, ``Minimizing conflicts:
  a heuristic repair method for constraint satisfaction and scheduling
  problems,'' {\em Artificial intelligence}, vol.~58, no.~1-3, pp.~161--205,
  1992.

\end{thebibliography}
\end{document}